\documentclass[journal]{IEEEtran}
\ifCLASSINFOpdf
\else
\fi
\usepackage{amsmath}
\usepackage{url} 

\usepackage{graphicx}
\usepackage{tabularx}
\usepackage{booktabs}     
\usepackage{amssymb}      
\usepackage{xcolor}       
\usepackage{graphicx}     
\usepackage{caption}      
\usepackage{subcaption} 
\usepackage{multirow} 
\begin{document}
%
\title{


Real-Time Human-Robot Interaction Intent Detection Using RGB-based Pose and Emotion Cues with Cross-Camera Model Generalization
}
%
%
%

\author{Farida Mohsen, Ali Safa,~\IEEEmembership{Member,~IEEE}
\thanks{F. Mohsen and A. Safa are with the College of Science and Engineering, Hamad Bin Khalifa University, Doha, Qatar. E-mail: fmohsen@hbku.edu.qa, asafa@hbku.edu.qa.}
\thanks{F. Mohsen carried the technical developments, the data collection and the design of the experiments. A. Safa supervised the project as Principal Investigator and contributed to the technical and experimental planning. All authors contributed to the writing of the manuscript.}
}

%
%

\markboth{Journal of \LaTeX\ Class Files,~Vol.~14, No.~8, August~2015}%
{Shell \MakeLowercase{\textit{et al.}}: Bare Demo of IEEEtran.cls for IEEE Journals}
%



\maketitle


\begin{abstract}

Service robots in public spaces require real-time understanding of human behavioral intentions for natural interaction. We present a practical multimodal framework for frame-accurate human–robot interaction intent detection that fuses camera-invariant 2D skeletal pose and facial emotion features extracted from monocular RGB video. Unlike prior methods requiring RGB-D sensors or GPU acceleration, our approach resource-constrained embedded hardware (Raspberry Pi 5, CPU-only). To address the severe class imbalance in natural human–robot interaction datasets, we introduce a novel approach to synthesize temporally coherent pose-emotion-label sequences for data re-balancing called MINT-RVAE (Multimodal Recurrent Variational Autoencoder for Intent Sequence Generation). Comprehensive offline evaluations under cross-subject and cross-scene protocols demonstrate strong generalization performance, achieving frame- and sequence-level AUROC of 0.95. Crucially, we validate real-world generalization through cross-camera evaluation on the MIRA robot head, which employs a different onboard RGB sensor and operates in uncontrolled environments not represented in the training data. Despite this domain shift, the deployed system achieves 91\% accuracy and 100\% recall across 32 live interaction trials. The close correspondence between offline and deployed performance confirms the cross-sensor and cross-environment robustness of the proposed multimodal approach, highlighting its suitability for ubiquitous multimedia-enabled social robots.

\end{abstract}


%
\IEEEpeerreviewmaketitle

\section*{Suplementary Material}
The supplementary materials associated with this article are available at: \texttt{https://tinyurl.com/374t2aw9}

\section{Introduction}
    
    
    

\begin{figure}[t]
    \centering
     \centering
    \includegraphics[width=0.49\textwidth]{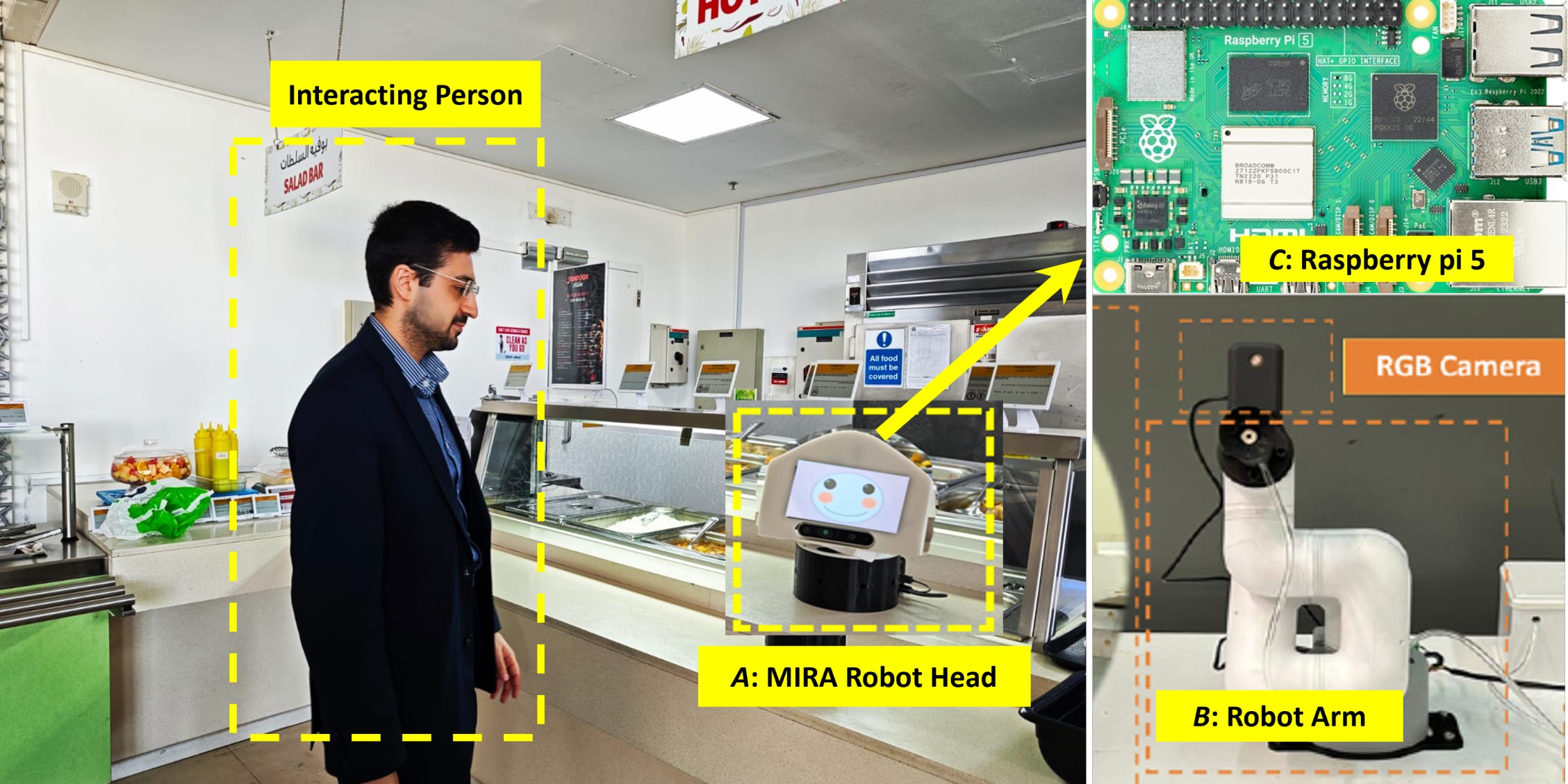}
    \caption{\textit{\textbf{Proposed human interaction intent detection setup.} The proposed system operates on two different platforms: (\textit{A}) the \textit{MIRA} robot, which integrates an expressive robotic head for HRI and an RGB camera for visual perception; and (\textit{B}) a MyCobot 320 setup equipped with a different RGB camera. Two-dimensional human pose features and facial emotion features are respectively extracted from the camera data using a YOLOv8-Pose model \cite{ultralytics_yolov8} and a DeepFace model~\cite{serengil2020deepface}. Then, the pose and emotion cues are fused into multimodal representations to predict interaction intent using various temporal models (GRU, LSTM, and a lightweight Transformer). (C) The proposed system runs in real-time on a resource-constrained embedded computing platform (Raspberry Pi 5) without GPU acceleration (CPU-only).} }
     \vspace{-2em}
    \label{fig:graphabstr}
\end{figure}

\IEEEPARstart{R}{obots} designed for social and service environments must not only perceive their surroundings but also infer the intentions of nearby humans to enable smooth, anticipatory, and safe interaction. The ability to recognize when a person intention to interact before any explicit contact forms a critical component of natural human–robot interaction (HRI) \cite{breazeal2003emotion, goodrich2008human, foster2021survey}. Early intention detection allows social robots to proactively adjust its behavior, initiate communication, or prepare for collaborative action, resulting in more fluid and human-compatible interactions \cite{lee2021service, bartneck2024human, yang2022model, valls2024robot}.  Furthermore, it is important to develop solutions that can generalize across various applications, allowing for rapid adaptation to specific tasks \cite{mavsar2023gan}. However, intention cues are subtle, context-dependent, and often ambiguous, typically arising from complex spatio–temporal combinations of motion and affective signals.

There has been substantial progress in computer vision and deep learning for recognizing human actions and intentions \cite{abbate2024self, Arreghini2024, trick2023can, liu2019rgbd, Xu2020HOI}, however, the development of robust intent-detection systems for affordable, low-power service robots remains a major challenge. Existing approaches often rely on RGB–D (depth) cameras \cite{zimmermann2018pose, abbate2024self}, which capture 3D human pose and spatial positioning to estimate approach trajectories and proxemic dynamics. However, the use of RGB–D sensors introduces several constraints for budget-conscious robotic platforms: \textit{(i)} \textit{hardware cost}, as commercial depth cameras (e.g., Intel RealSense, Microsoft Kinect) typically cost  $\$350–600$ compared to $\$10–30$ for standard RGB webcams, limiting scalability; \textit{(ii)} \textit{computational demands}, as depth-based methods generally require desktop-class GPUs \cite{liu2019rgbd}, impractical for resource-constrained social robots; and \textit{(iii)} \textit{calibration overhead}, as RGB–D systems often necessitate per-device calibration and exhibit poor generalization across camera hardware \cite{zhang2025comprehensive}. Although alternative sensing modalities such as wearable inertial units \cite{nshimyimana2025pim}, tactile sensors \cite{Wong2022VisionTactile} and eye-trackers \cite{belcamino2024gaze} have been explored for human-intention recognition, skeleton-based pose sequences extracted from standard RGB cameras offer a more cost-effective, computationally efficient, and privacy-preserving solution for real-world interaction-intent detection \cite{zolfaghari2024sensor}.

Crucially, the field still lacks a systematic investigation into whether \textit{RGB-only vision models, when executed on embedded platforms without GPU acceleration, can achieve competitive performance for service-robot intent detection \cite{Wang2024RTHARE}.} This constitutes the first major limitation addressed in this work. In parallel, most prior studies have evaluated intent recognition only under \textit{controlled laboratory settings} with scripted participant behaviors \cite{zimmermann2018pose, trick2023can, ngo2024toward, 10582024, 9537632}, leaving open the question of how such systems perform in naturalistic, real-world environments. A further methodological limitation concerns the temporal resolution of existing intent detectors: most systems report only \textit{sequence-level} or \textit{event-level} classifications with replicated frame labels \cite{Arreghini2024,abbate2024self,trick2023can, Kedia2024}, which undermines early intention detection. Another challenge is that real-world HRI data further exhibit pronounced class imbalance: interaction episodes are rare in real deployments (e.g., 112 interacting vs. 4,245 non-interacting trajectories in PAR-D \cite{thompson2024pard}, and engagement rates of \(\sim 3.6\%\) in mall studies \cite{natori2025mall}). This imbalance occurs at both sequence and frame levels, and common remedies such as undersampling \cite{del2020still} or SMOTE-like oversampling \cite{chawla2002smote} are poorly suited to sequential multimodal data because they reduce negative-data diversity or disrupt temporal and cross-modal coherence. These limitations motivate learning-based generative re-balancing, such as VAEs \cite{kingma2014autoencoding} or GANs \cite{goodfellow2020generative}, to synthesize temporally coherent and semantically aligned sequences.

To address these limitations, we present a practical and deployable framework for real-time human interaction intent detection using only monocular RGB input, running entirely on a resource-constrained CPU-only embedded platform (Raspberry Pi 5, see Fig. \ref{fig:graphabstr}). The presented solution is capable of operating in real time on resource-constrained robotic platforms, with explicit handling of class imbalance and validation across multiple real-world scenarios. Our proposed system jointly integrates camera-invariant 2D body-pose and facial-emotion features within lightweight temporal models including \textit{Gated Recurrent Unit} (\texttt{GRU}) \cite{Chung2014GRUEval}, \textit{Long Short-Term Memory} (\texttt{LSTM}) \cite{hochreiter1997lstm}, and \texttt{Transformer} \cite{vaswani2017attention} to infer the onset of human interaction intent without requiring depth sensors or subject identity recognition. We demonstrate that our proposed pre-processing approach based on the bounding-box normalization of skeletal keypoints yields camera-invariant features that \textit{generalize} well across multiple hardware and mounting configurations \textit{without model fine-tuning}. This enables training on commodity USB webcams and later deployment on different robot-mounted cameras. To mitigate severe class imbalance, a central aspect of this work is our newly proposed \textit{(Multimodal Recurrent Variational
Autoencoder for Intent Sequence Generation} (MINT-RVAE), which consists of a generative model that synthesizes temporally coherent pose–emotion sequences aligned with intention labels to augment rare intent-onset events, thereby enhancing the generalization and discriminative capacity of detection models. This work has the following key contributions:
\begin{enumerate}

\item \emph{RGB-only multimodal intent detection framework}: We propose a deployable system combining camera-invariant 2D body pose and facial emotion features within temporal modeling architectures (GRU, LSTM, Transformer) for frame-level intent prediction from monocular RGB video. Unlike prior work relying on RGB-D sensors or GPU acceleration, our approach enables deployment on resource-constrained embedded platforms.
\item \emph{MINT-RVAE for HRI data re-balancing}: We introduce a Multimodal Imbalance-Aware Recurrent Variational Autoencoder that generates temporally coherent, cross-modal pose–emotion sequences aligned with intention labels, addressing the severe class imbalance inherent in natural HRI datasets.

\item \emph{Comprehensive cross-domain and cross-sensor validation:}
We perform rigorous evaluation across three complementary protocols including  cross-subject (5-fold cross-validation (CV)), cross-scene (held-out multi-person environment) and cross-camera deployment to assess both generalization and deployability. Our model achieves 0.95 AUROC in offline 5-fold cross-validation and 91\% accuracy in real-world trials on the MIRA robot head, running entirely on a CPU-only Raspberry Pi 5. Notably, the MIRA platform employs a different onboard RGB sensor and operates in uncontrolled environments absent from the training data, yet exhibits comparable performance to offline evaluation. This consistency demonstrates the proposed system’s strong cross-camera and cross-environment generalization under realistic operating conditions.

\end{enumerate}

The remainder of this paper is organized as follows. Section~\ref{sec:related} reviews related work. Section~\ref{sec:approach} describes the data collection setup and  proposed RGB-only perception framework. Section~\ref{mintrvae} introduces our novel MINT-RVAE framework, detailing its architecture, training objectives, and validation of synthetic sequence quality. Section~\ref{sec:results} presents comprehensive experimental results. Finally, Section~\ref{sec:conclusion} summarizes the main findings and outlines directions for future research.


\section{Related Works}
\label{sec:related}
HRI has garnered extensive attention in recent years due to the growing demands of service robots in both domestic and industrial contexts~\cite{su2023recent}, where autonomous systems must collaborate effectively with humans to accomplish shared tasks. The overarching research goals in HRI include enhancing task efficiency, enabling robot learning through physical interaction, and achieving smooth, natural collaboration~\cite{asif2026exploring}.
Beyond HRI, intention modeling has also been explored in the broader multimedia domain, where visual cues from human–object interactions are used to infer latent goals and intentions from RGB video~\cite{Xu2020HOI}. Building on these foundations, robots in human-centered environments require perceptual interfaces capable of anticipating human actions and intentions, making intention recognition a cornerstone capability for cooperative HRI.

\textit{Sensor Modalities: } A wide range of sensors have been explored for recognizing human intent in HRI. Many state-of-the-art systems leverage \emph{multimodal} inputs beyond a single RGB camera. Depth sensors (RGB-D) are are particularly prevalent, as they provide 3D skeletal tracking to improve intent recognition accuracy \cite{Kedia2024,Arreghini2024}. For example, recent intent-prediction models for human–robot collaboration often assume an RGB-D camera or motion-capture system to capture human pose \cite{Kedia2024,Arreghini2024}, or even wearable inertial units in industrial environments \cite{Zhang2023Review}. While these configurations enhance precision, they substantially increase cost, complexity, and deployment effort \cite{Gaschler2012,Belardinelli2022}. Other works incorporate complementary modalities such as eye gaze and audio. Belardinelli \emph{et al.} \cite{Belardinelli2022} combine a gaze-tracking interface with hand-motion features to infer the operator’s intention in a shared-control task, while Arreghini \emph{et al.} \cite{Arreghini2024} demonstrate that including user gaze cues (e.g. head pose or eye contact) significantly boosts detection of a person’s intent to engage a service robot. Similarly, voice-based interaction cues (e.g., wake words such as “Hey robot”) are commonly integrated in multimodal frameworks, and several reviews emphasize that combining visual, auditory, and gaze-based information is now a dominant trend in intention-aware HRI design \cite{Zhang2023Review}. In contrast to these hardware-intensive approaches, our framework relies solely on a monocular RGB camera. By extracting multiple visual cues including 2D body pose and facial emotion, it captures the rich nonverbal signals (posture, proxemics, affect) known to convey engagement intent \cite{Gaschler2012,Arreghini2024}.
This design avoids depth sensors or on-body devices, resulting in a cost-effective, calibration-free system suitable for real-world social robots.

\textit{Class Imbalance in HRI Data:} A key challenge for intent detection is the severe class imbalance inherent in naturalistic HRI data. Engagement or interaction events are relatively rare compared to prolonged periods of non-interaction. Many prior HRI datasets contain overwhelmingly more negative/no-intent frames than positive intent frames \cite{trick2023can,Thompson2024}. Without countermeasures, this imbalance can lead learning algorithms to bias toward the majority class (predicting “no intent” for everything). Despite this, few studies explicitly address imbalance beyond simple oversampling or class-weighting. For instance, a common approach is to synthetically oversample scarce events using techniques like SMOTE \cite{Chawla2002}, or to record additional positive examples through staged interactions \cite{Thompson2024}. These strategies help but do not fully capture the temporal context of intent cues. In this work, we address imbalance at the sequence level through a learning-based generative augmentation strategy.   We introduce (MINT-RVAE), which \emph{generates new  coherent synthetic sequences}  to enrich the minority class. This approach differs from standard oversampling \cite{Chawla2002} by creating novel pose–emotion trajectories, not just reweighting or duplicating existing samples. To our knowledge, only few recent works have begun to explore such learning-based augmentation for HRI. Abbate \emph{et al.} \cite{abbate2024self} propose a self-supervised pre-training scheme to exploit unlabeled human approach data, which increases the effective amount of positive examples without manual annotation. Our solution instead explicitly models minority-class sequences and balances the training set by augmenting under-represented intent segments.   This targeted rebalancing significantly enhances model sensitivity to rare behaviors, addressing one of the most persistent limitations in HRI intent recognition.


\begin{figure*}[t]
    \centering
    \includegraphics[width=0.95\textwidth]{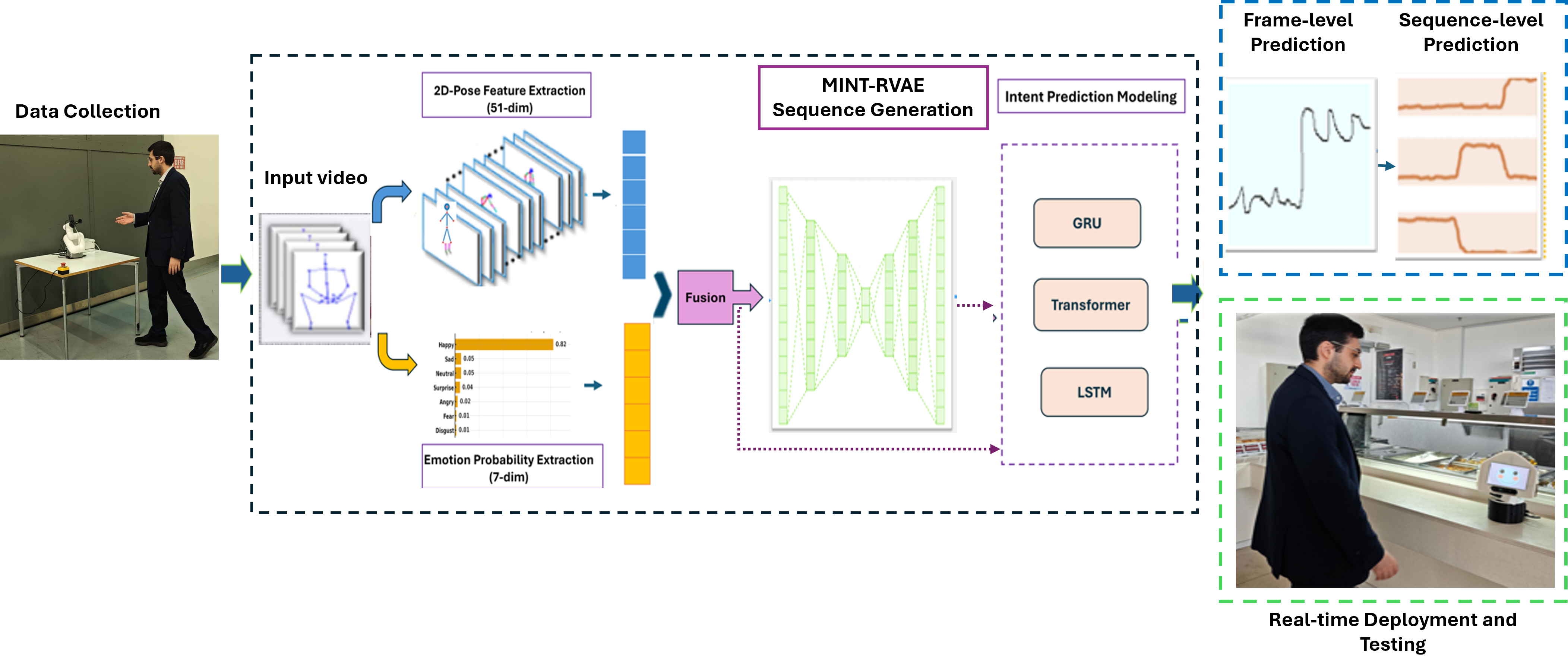}
  \caption{\textit{\textbf{Proposed data processing pipeline.} The RGB camera feed from the robot arm is processed using a \texttt{YOLOv8-pose} model for pose coordinates and \texttt{DeepFace} for emotion vectors. These outputs are concatenated into multimodal feature vectors, which serve as input to various intent detection backbones studied in this work (\texttt{GRU}, \texttt{LSTM}, \texttt{Transformer}). To address the class imbalance in HRI data, our proposed \textit{MINT-RVAE} generates synthetic sequences during training. The pipeline supports both frame-level and sequence-level intent prediction. 
  } }
   \label{fig:study_diagram}
   \vspace{-1em}
\end{figure*}

 \textit{Temporal Modeling and Intent Granularity:} 
 Human intent develops dynamically, making temporal modeling a core component of intent recognition. Early studies employed recurrent neural networks (RNNs) to capture motion dynamics over time.
Yu and Lee \cite{Yu2015Dynamic} demonstrated that hierarchical multi-timescale RNNs could anticipate human intent (e.g., approach or gesture) from continuous movement patterns, laying the foundation for temporal intent prediction.

Subsequent works widely adopted gated RNN architectures such as LSTMs and GRUs for sequence learning.
For instance, Trick \textit{et al.} \cite{trick2023can} used an LSTM classifier to detect help-seeking behavior by aggregating short video segments, while several other studies adopted similar window-based approaches that classify short clips (typically 2–3~s) as either “intent” or “no intent” \cite{Arreghini2024,abbate2024self}.
Although these methods capture temporal dependencies, their segment-level inference inherently limits temporal precision, since a single label is assigned to an entire clip rather than pinpointing the onset of intent \cite{Arreghini2024,abbate2024self}.
Some approaches even perform labeling post-hoc—only after detecting the full interaction—preventing proactive robot behavior \cite{trick2023can}. In contrast, our framework performs frame-wise intent inference, continuously processing multimodal pose–emotion streams and estimating an intent probability for each frame. This fine-grained temporal resolution allows early and proactive intent detection, enabling a robot to initiate engagement precisely when intent emerges rather than after the interaction happens.

\textit{Real-Time Embedded Deployment:} 
The feasibility of deploying intent detection models on embedded platforms is critical for practical HRI, yet most existing studies overlook these resource constraints. Many recent approaches employ deep architectures (e.g., large CNNs or Transformers) and report results using GPU-accelerated inference on offline datasets \cite{Kedia2024,Arreghini2024,abbate2024self}.
For instance, Kedia \textit{et al.} \cite{Kedia2024} trained Transformer-based intent predictors for human–robot manipulation tasks, but their approach assumes access to high-performance computing resources and does not consider on-board computation.
Similarly, Arreghini \textit{et al.} \cite{Arreghini2024} achieved strong intent classification performance (AUROC~$>$~0.85) by fusing video and gaze cues, yet their evaluation was conducted on workstation GPUs rather than embedded devices. Overall, the majority of intent-recognition frameworks emphasize accuracy over deployability, with limited consideration for real-time, resource-constrained scenarios.

A notable exception is the early work by Gaschler \textit{et al.} \cite{Gaschler2012}, who implemented a rule-based posture analysis system on a robot using a Kinect sensor for real-time interaction. However, the simplicity of that approach restricted its generalization capability across different contexts.
In contrast, our proposed system is designed with computational efficiency as a first-class objective. By combining lightweight temporal models with optimized feature extraction, it operates in real time on a CPU-only embedded platform (Raspberry~Pi~5) without GPU acceleration. To the best of our knowledge, this constitutes one of the first HRI intent-detection frameworks experimentally validated directly on embedded hardware. 

To summarize, in comparison to prior art our proposed system offers: (i) purely monocular vision-based sensing (vs. depth or wearable sensors) with multimodal cue fusion, (ii) true frame-level intent onset detection (vs. segment-level predictions), (iii) an imbalance-aware training strategy via synthetic sequence generation, and (iv) validated operation on low-power hardware in real time. These distinctions mark a significant step toward practical and generalizable intent recognition for human–robot interaction.

\section{Data Collection Setup and Pre-Processing}
\label{sec:approach}
We address the problem of recognizing human intention to interact with a robot operating in shared spaces using only monocular RGB input. The robot is assumed to observe nearby individuals through an onboard camera and must distinguish between \textit{passersby} and those intending to \textit{engage}, enabling anticipatory and context-aware behavior. Our framework aims to detect this intention as early and accurately as possible from purely visual cues. The proposed system learns discriminative spatio–temporal patterns from multimodal information extracted from RGB video. Specifically, it integrates two complementary perceptual channels: \textit{body-pose dynamics}, capturing approach and gesture cues, and \textit{}{facial affect features}, capturing emotional status. These features are combined into compact multimodal representations that are processed by lightweight temporal models including \emph{GRU}, \emph{LSTM}, and  a compact \emph{Transformer} to predict interaction intent at both frame and sequence levels. Unlike standard Transformer architectures, our Transformer variant comprises only a single encoder block with learnable positional embeddings and four attention heads, ensuring computational efficiency suitable for embedded deployment. As illustrated in Fig.~\ref{fig:study_diagram}, the end-to-end pipeline comprises (i) RGB-based perception for human detection and feature extraction, (ii) \emph{MINT-RVAE} generative model synthesizes realistic pose–emotion sequences consistent with intent labels, improving data diversity and recall for rare interaction events, (iii)  multimodal temporal intent modeling, and (iii) deployment on a real robot platform in uncontrolled real-world environment. 

\subsection{Robot Hardware}

\begin{figure}[t]
    \centering
    \begin{subfigure}[t]{0.4\linewidth}
        \centering
        \includegraphics[width=\linewidth]{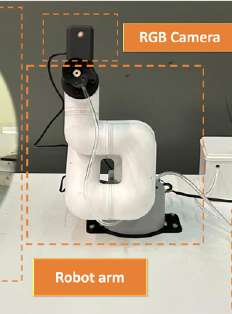}
         \caption{}
    \end{subfigure}
    \hspace{0.001\linewidth}
    \begin{subfigure}[t]{0.53\linewidth}
        \centering
        \includegraphics[width=\linewidth]{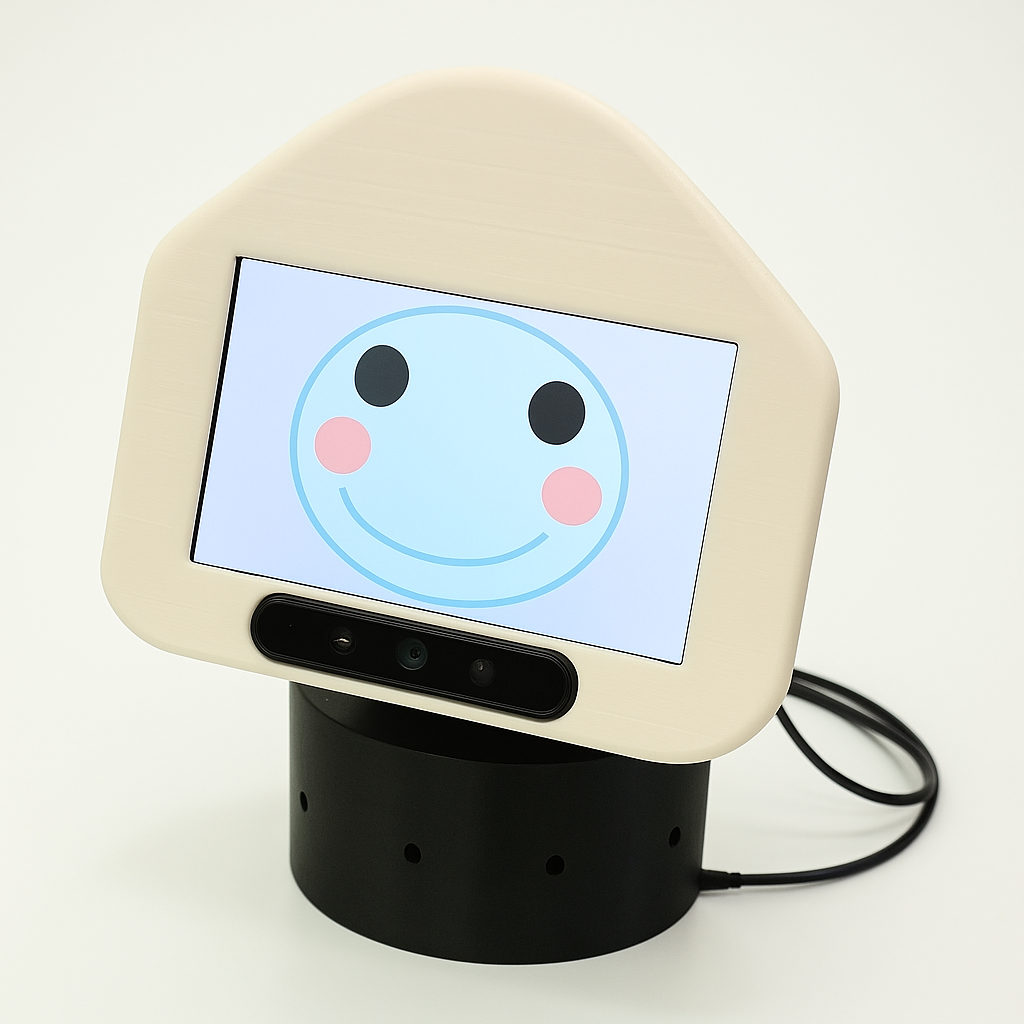}
         \caption{}
    \end{subfigure}
\caption{\textit{\textbf{Experimental robotic platforms.} (a) Data collection setup using an Elephant Robotics MyCobot 320 arm equipped with a monocular RGB camera for recording human–robot approach sequences. (b) MIRA social robot head used for real-time deployment experiments embedded with Raspberry Pi 5 that performs CPU-only inference with onboard RGB sensing during evaluation.}}
    \vspace{-2em}
    \label{fig:robots}
\end{figure}

Two robotic platforms serve complementary roles in our study 
(Fig.~\ref{fig:robots}). During dataset acquisition, we employ an 
Elephant Robotics MyCobot 320 collaborative arm equipped with 
a standard USB webcam. This  configuration provides flexibility for camera positioning and controlled 
recording of human approach behaviors. Participants annotate interaction intent at the frame level using a 
wireless presenter: pressing the button at the moment of engagement 
decision generates a timestamped onset signal synchronized with the 
video stream.

For real-world deployment validation, we integrate the trained model into a MIRA social robot head 
(Fig.~\ref{fig:robots}b). This robot head has a limited compute power using a Raspberry~Pi~5 (8\,GB RAM, ARM Cortex-A76 at 2.4\,GHz) providing \textit{mobile-grade} CPU-only execution (no \textit{desktop-grade} CPU nor GPU acceleration), representative of cost-sensitive service robots deployed in public spaces. Crucially, 
the camera sensor and image properties differ substantially from 
the USB training setup, enabling rigorous evaluation of cross-camera 
generalization ability of our proposed approach during the experiments.

\subsection{Dataset}
\label{dataset}
\begin{figure}[]
    \centering
    \begin{subfigure}[t]{0.5\linewidth}
        \centering
        \includegraphics[width=0.9\linewidth]{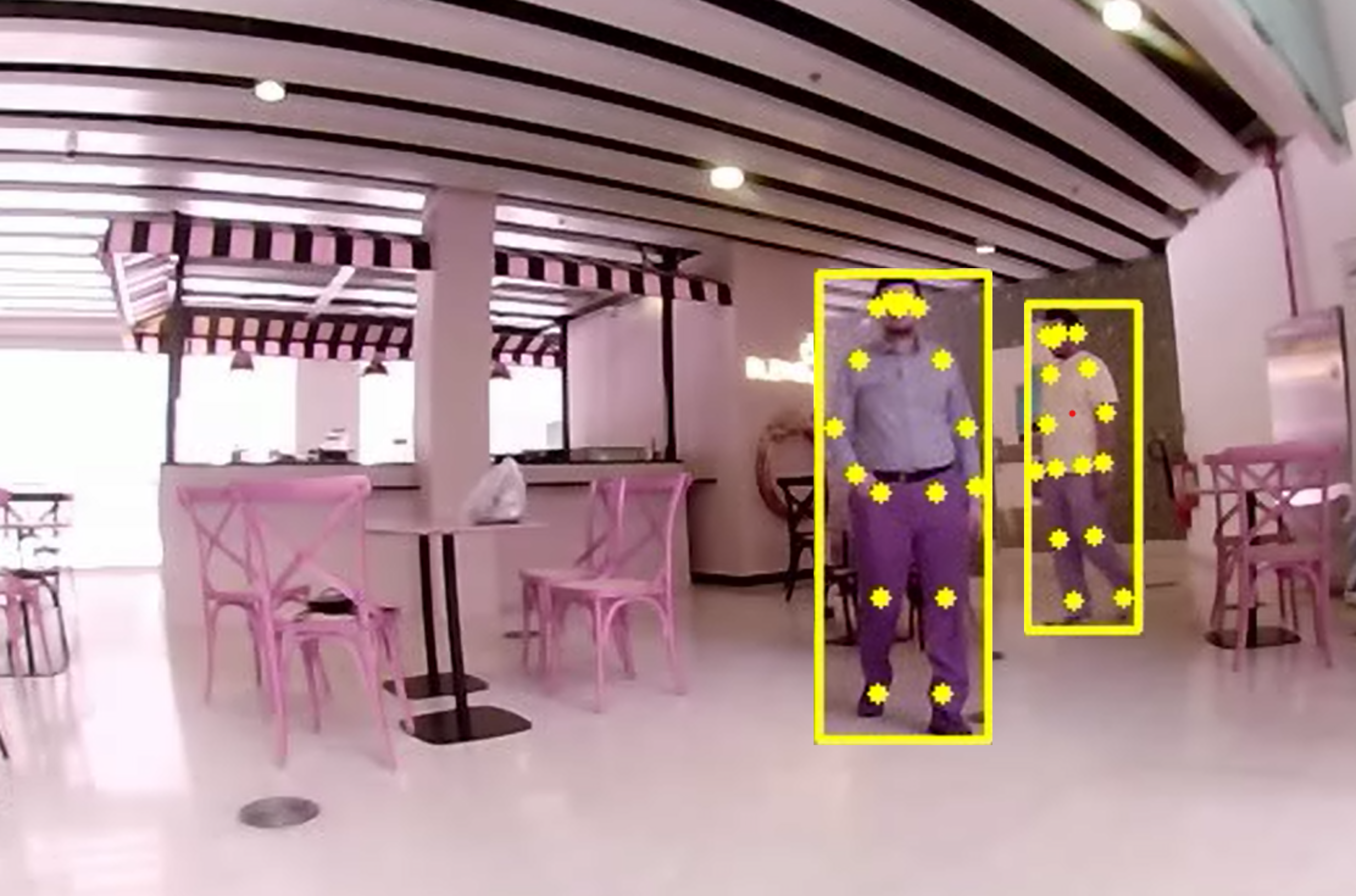}
        \label{fig:one_person}
    \end{subfigure}
    \hfill
    \begin{subfigure}[t]{0.48\linewidth}
        \centering
        \includegraphics[width=0.9\linewidth]{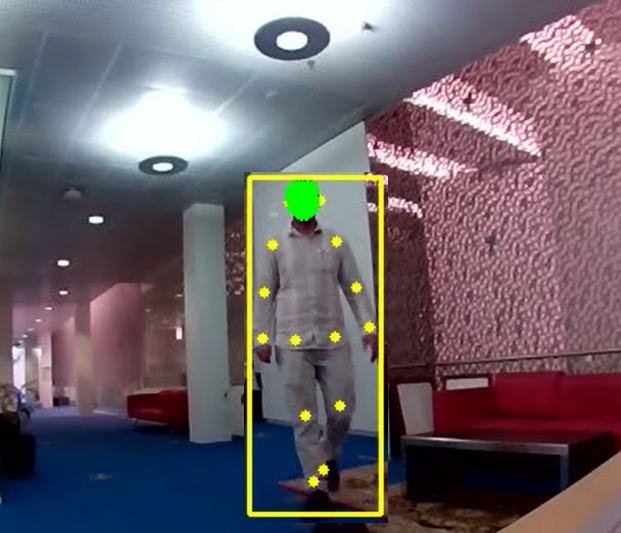}
        \label{fig:2_person}
    \end{subfigure}
\caption{\textit{\textbf{Views from our data collection.}}}
    \label{fig:snapshoots}
    \vspace{-2em}
\end{figure}

We collected RGB video sequences capturing human–robot  approach behaviors with $10$ different participants in three indoor environments with varying characteristics. Environment~1 (library study area) contains 54~sequences 
(7,620~frames, 30.2\% intent) with controlled lighting and no  background motion. Environment~2 (building corridor) comprises 23~sequences (3,900~frames, 32.8\% intent) introducing minimal pedestrian 
 and dynamic lighting. Environment~3 (public cafeteria) 
features 11~sequences (1,095~frames, 37.5\% intent) with multi-person interactions, visual clutter, and unstructured scene geometry. Fig.~\ref{fig:snapshoots} shows representative frames from each setting.

The dataset exhibits class imbalance characteristic of real-world HRI 
deployments, where prolonged non-interaction periods dominate observation 
windows \cite{thompson2024pard, natori2025mall}. Approximately 
70\% of frames correspond to no-intent states (passersby, ambient motion), 
while 30\% capture approach and engagement behaviors. This distribution 
reflects naturalistic service-robot operation and motivates our 
imbalance-aware augmentation strategy (Section~\ref{mintrvae}).

Frame-level annotations were obtained via synchronized button presses: 
participants held a wireless presenter and activated it at the precise 
moment they formed the intention to interact, generating timestamps 
aligned with video frame indices. All participants agreed to participate in the data collection. To preserve privacy, we do not retain or distribute raw video; instead, we extract only de-identified 2D pose coordinates and emotion probability vectors as described below. 

The dataset is organized into a collection of sequences. Let $Q$ denote the total number of recorded sequences. Sequence $r \in \{1,\dots,Q\}$ contains frames indexed by $i=1,\ldots,T_r$ with associated feature vectors $\mathbf{z}_{i,r}$ and binary frame-level intent labels $\ell_{i,r}\in\{0,1\}$, where $\ell_{i,r}=1$ indicates \emph{intent} and $\ell_{i,r}=0$ indicates \emph{no-intent}. 
\begin{equation}
D=\{(\mathbf{z}_{i,r},\ell_{i,r})\mid r=1,\dots,Q;\; i=1,\dots,T_r\}.
\label{eq:data_repr}
\end{equation}
Each feature vector concatenates pose and emotion components $\mathbf{g}^{\text{pose}}_{i,r}, \mathbf{g}^{\text{emo}}_{i,r}$ extracted from the camera data:
\begin{equation}
    \mathbf{z}_{i,r}=\big[\mathbf{g}^{\text{pose}}_{i,r};\, \mathbf{g}^{\text{emo}}_{i,r}\big].
\end{equation}



\subsection{Multimodal Feature Extraction}
\label{subsec:prepro}

In this study, we leverage proxemics i.e. the analysis of human motion and spatial behavior—as a primary cue for inferring interaction intent. In particular, proxemics examines how individuals occupy and navigate the robot’s social space, providing valuable indicators of engagement likelihood~\cite{saunderson2019robots}. Such motion-based cues are especially relevant for early intention recognition, as they naturally precede verbal or physical interaction. Accordingly, two complementary perceptual channels are extracted from the monocular RGB stream: body-pose keypoints and facial emotion features:

\subsubsection{Camera-Invariant 2D Pose Features}  
The first perceptual level captures human body posture and motion dynamics through 2D skeletal keypoints. These features are extracted frame-by-frame using the \texttt{YOLOv8-pose} model~\cite{ultralytics_yolov8}, which provides pixel coordinates $(u_n,v_n)$ and detection confidences $s_n$ for each keypoint $n=1,\dots,M$, together with a person bounding box $\mathcal{B}_{i,r}=(u_{\min},v_{\min},w,h)$. To ensure scale and translation invariance across scenes, keypoints are normalized within the bounding box:

\begin{equation}
(\tilde{u}_n,\tilde{v}_n)=\Big(\tfrac{u_n-u_{\min}}{w},\,\tfrac{v_n-v_{\min}}{h}\Big).
\end{equation}
The pose descriptor is then formed as:
\begin{equation}
\mathbf{g}^{\text{pose}}_{i,r}=\big[\,\tilde{u}_1,\tilde{v}_1,s_1,\,\dots,\,\tilde{u}_M,\tilde{v}_M,s_M\,\big]\in\mathbb{R}^{3M},
\end{equation}
where $(\tilde{u}_n,\tilde{v}_n)$ are standardized using training-split statistics (mean and standard deviation), while confidence scores $s_n$ remain unaltered. 

\subsubsection{Facial Emotion Features}
The second perceptual level focuses on affective expressions, which provide complementary cues to body pose. For each detected person, facial regions are identified using the same \texttt{YOLOv8} detector~\cite{ultralytics_yolov8}, cropped, and analyzed by \texttt{DeepFace}~\cite{serengil2020deepface} to estimate emotion probabilities. Each face yields a seven-dimensional probability vector over the basic emotions—happy, sad, neutral, surprise, angry, fear, and disgust—denoted as:
\begin{equation}
\mathbf{g}^{\text{emo}}_{i,r}=[q_1,\dots,q_7], \quad q_m \in [0,1], \;\; \sum_{m=1}^{7} q_m = 1.
\end{equation}
These emotion probabilities are temporally aligned with the corresponding pose features to form a unified multimodal representation, later fused and processed by the temporal intention-prediction models described in Section~\ref{seq_models}.


\section{Multimodal Recurrent Variational Autoencoder for Intent Sequence Re-Balancing (MINT-RVAE)}

\label{mintrvae}

\begin{figure*}[t]
    \centering
    \includegraphics[width=0.99\linewidth]{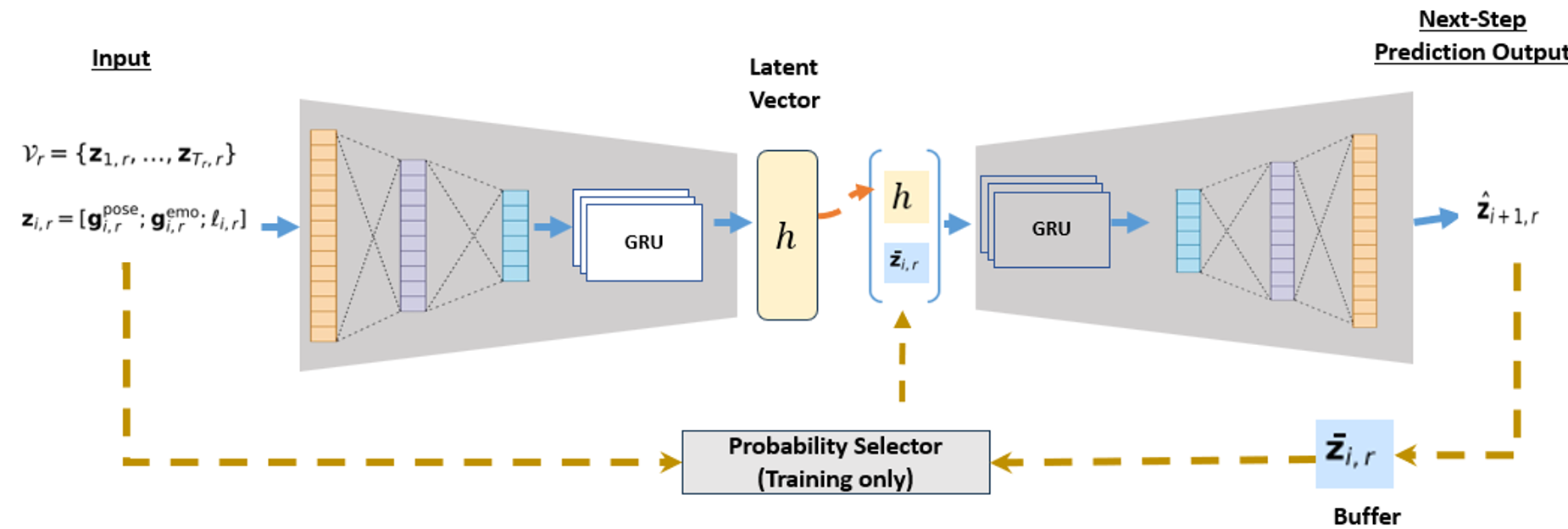}
\caption{\textit{\textbf{The proposed MINT-RVAE architicture.} 
An input sequence $\mathcal{V}_r = \{\mathbf{z}_{1,r}, \dots, \mathbf{z}_{T_r,r}\}$ (each frame concatenating $\mathbf{g}^{\text{pose}}_{i,r}$, $\mathbf{g}^{\text{emo}}_{i,r}$, and $\ell_{i,r}$) is first processed by an MLP before being encoded by a GRU. 
The final encoder state parameterizes the latent vector $\mathbf{h} \in \mathbb{R}^{32}$ using the reparameterization trick. 
During decoding, $\mathbf{h}$ initializes the decoder GRU hidden state and is concatenated with the previous input $\tilde{\mathbf{z}}_{i,r}$ at each time step to predict the next multimodal frame $\hat{\mathbf{z}}_{i+1,r}$. 
The \textit{Probability Selector} block (used only during training) controls the teacher-forcing rate, determining whether the decoder receives the ground-truth or predicted input. 
During inference, the latent vector $\mathbf{h}$ is sampled from the standard normal prior $\mathcal{N}(\mathbf{0}, \mathbf{I})$, and the decoder operates in a fully autoregressive mode, feeding back its own predictions $\hat{\mathbf{z}}_{i,r}$ as input.}}
\label{fig:rvae_schematic}

\vspace{-1em}
\end{figure*}

The collected dataset exhibits a substantial imbalance between \emph{intent} and \emph{no-intent} samples, with only approximately $30\%$ of frames labeled as \emph{intent} as described in Section \ref{dataset}. To mitigate this, we introduce the \emph{MINT-RVAE} (Fig.~\ref{fig:rvae_schematic}), a generative model that produces synthetic pose–emotion–label sequences that aims to mimic the dynamics of real interactions, thereby improving minority-class representation. Training the proposed \textit{MINT-RVAE} presents unique challenges due to the \textit{heterogeneous} nature of its multimodal inputs, where pose and emotion channels contribute information with differing \textit{scales}, \textit{distributions}, and \textit{semantic relevance}. Without careful design, the model may overfit to a single modality, compromising its ability to generate truly \textit{balanced multimodal} sequences. Moreover, standard reconstruction objectives commonly used in \textit{VAEs} (e.g., mean squared error) proved inadequate for accurately reconstructing the structured, low-dimensional \textit{pose–emotion representations} employed in this work. To overcome these limitations, we introduce \textit{custom loss formulations} that jointly preserve \textit{temporal coherence} and \textit{reconstruction fidelity}.

\subsection{MINT-RVAE Architecture}

\subsubsection{Input Sequences}
The proposed \textit{MINT-RVAE} (see Fig.~\ref{fig:rvae_schematic}) is designed to jointly reconstruct pose dynamics, facial-affect evolution, and associated intent labels from  multimodal temporal segments. Each input sequence $r \in \{1, \ldots, Q\}$ is composed of $T_r$ frames, each represented by a multimodal feature vector $\mathbf{z}_{i,r}$ and its corresponding binary label $\ell_{i,r} \in \{0,1\}$. For sequence $r$, we define:
\begin{equation}
\mathcal{V}_r = \{\mathbf{z}_{1,r}, \mathbf{z}_{2,r}, \dots, \mathbf{z}_{T_r,r}\}, 
\qquad \mathbf{z}_{i,r} \in \mathbb{R}^D,
\label{eq:mint_input}
\end{equation}
where each feature vector $\mathbf{z}_{i,r} = [\mathbf{g}^{\text{pose}}_{i,r};\, \mathbf{g}^{\text{emo}}_{i,r};\, \ell_{i,r}]$ concatenates:  
\textit{(i)} a 2D body-pose descriptor $\mathbf{g}^{\text{pose}}_{i,r} \in \mathbb{R}^{3M}$;  
\textit{(ii)} a 7-dimensional emotion-probability vector $\mathbf{g}^{\text{emo}}_{i,r} \in \mathbb{R}^7$; and  
\textit{(iii)} the frame-level interaction label $\ell_{i,r}$.  
This yields a multimodal feature dimensionality of $D = 3M + 8 + 1 = 59$ (with $M = 17$ keypoints). 

\subsubsection{Encoder}
The encoder $q_{\phi}(\mathbf{h}|\mathcal{V}_r)$ maps an input sequence $\mathcal{V}_r$ into a latent representation distributed according to a standard Gaussian  prior.  Each frame-level feature vector $\mathbf{z}_{i,r}$ is first passed through a three-layer Multilayer Perceptron (MLP) with layer dimensions $256 \!\to\! 128 \!\to\! 64$, each followed by batch normalization, ReLU activation, and dropout. This stage performs \textit{modality fusion} and dimensionality compression.  
The resulting sequence of embeddings is then processed by a \textit{GRU} which encodes temporal dependencies across consecutive frames.  Then, the GRU’s final hidden state is transformed through two parallel linear projections, $\mu_{\phi}(\cdot)$ and $\sigma_{\phi}(\cdot)$, yielding the mean and variance of the approximate posterior:
\begin{equation}
q_{\phi}(\mathbf{h}_r | \mathcal{V}_r) = 
\mathcal{N}\!\left(\mathbf{h}_r; \, \mu_{\phi}(\mathcal{V}_r), \,
\mathrm{diag}\!\left(\sigma_{\phi}^2(\mathcal{V}_r)\right)\right).
\end{equation}
A latent vector $\mathbf{h}_r \in \mathbb{R}^{32}$ is then sampled using the \textit{reparameterization trick}~\cite{kingma2014autoencoding}:
\begin{equation}
\mathbf{h}_r = \mu_{\phi}(\mathcal{V}_r) + 
\sigma_{\phi}(\mathcal{V}_r) \odot \boldsymbol{\epsilon}, 
\qquad \boldsymbol{\epsilon} \sim \mathcal{N}(\mathbf{0}, \mathbf{I}).
\end{equation}

\subsubsection{Decoder}
The decoder $p_{\theta}(\mathcal{V}_r | \mathbf{h}_r)$ reconstructs the input sequence in an \textit{autoregressive} manner.  
A GRU-based temporal decoder, initialized with the latent state $\mathbf{h}_r$, predicts the next-step multimodal vector $\hat{\mathbf{z}}_{i+1,r}$ conditioned on the previous hidden state $\mathbf{h}_{i}$ and the latent representation.  
The decoder output is mapped through an output MLP to reconstruct pose, emotion, and label components:
\begin{equation}
\hat{\mathbf{z}}_{i+1,r} = 
\mathrm{concat}(\hat{\mathbf{g}}^{\text{pose}}_{i+1,r},\,
\hat{\mathbf{g}}^{\text{emo}}_{i+1,r},\,
\hat{\ell}_{i+1,r}),
\end{equation}

where $\hat{\mathbf{g}}^{\text{pose}}_{i+1,r}$ is produced via a linear projection,  
$\hat{\mathbf{g}}^{\text{emo}}_{i+1,r}$ is normalized through a softmax layer,  
and $\hat{\ell}_{i+1,r}$ is obtained via a sigmoid activation to estimate the interaction probability.

\subsection{MINT-RVAE Training Approach and Loss Formulation}
\label{sec:mint_training}
For each temporal window, the model receives an input subsequence 
$\{\mathbf{z}_{1,r}, \ldots, \mathbf{z}_{T-1,r}\}$ and learns to predict the corresponding future frames 
$\{\mathbf{z}_{2,r}, \ldots, \mathbf{z}_{T,r}\}$.  
The encoder summarizes the multimodal input sequence into a latent code $\mathbf{z}_r$, which the decoder then uses to reconstruct the \textit{next-step} multimodal outputs in an autoregressive fashion.  
This next-frame prediction task enforces temporal consistency, encouraging the model to learn smooth and coherent generative dynamics across pose, emotion, and label modalities.


To enable coherent reconstruction of pose, affective, and intent dynamics, the proposed MINT-RVAE is optimized using a joint objective that combines multiple reconstruction terms with a latent regularization component. The overall objective function is expressed as:
\begin{equation}
\mathcal{J}_{\text{total}} 
= \alpha_p\,\mathcal{J}_{\text{pose}}
+ \alpha_e\,\mathcal{J}_{\text{emotion}}
+ \alpha_i\,\mathcal{J}_{\text{intent}}
+ \eta_z\,\mathcal{J}_{\text{KL}},
\label{eq:mint_total_loss_new}
\end{equation}
where $\mathcal{J}_{\text{pose}}$, $\mathcal{J}_{\text{emotion}}$, and $\mathcal{J}_{\text{intent}}$ denote reconstruction losses for 2D skeletal pose, facial emotion probabilities, and frame-level intent prediction, respectively. $\mathcal{J}_{\text{KL}}$ represents the variational regularization term that aligns the latent posterior with a unit Gaussian prior. The coefficients $\alpha_p$, $\alpha_e$, $\alpha_i$, and $\eta_z$ control the relative contribution of each objective term tuned empirically to preserve reconstruction fidelity while maintaining latent-space regularization.

\subsubsection{Pose Reconstruction Objective}
To achieve stable learning under partial occlusions and pose-estimation noise, a \textit{confidence-weighted smooth-$\ell_1$} objective is formulated by combining joint-coordinate and confidence-map reconstruction terms:
\begin{multline}
\mathcal{J}_{\text{pose}}
= \xi_1 \cdot \frac{1}{T_r-1}
\sum_{i=1}^{T_r-1}\sum_{n=1}^{M}
(\sigma_{i+1,n}+\epsilon_0)\,
\psi_\delta\!\big(\tilde{p}_{i+1,n}-p_{i+1,n}\big)
\\
+ \, \xi_2 \cdot 
\mathrm{MSE}(\tilde{\boldsymbol{\sigma}}_{2:T_r},\,\boldsymbol{\sigma}_{2:T_r}),
\label{eq:new_pose_loss}
\end{multline}
where $p_{i,n}$ and $\sigma_{i,n}$ denote the normalized 2D coordinates and detection confidence of joint $n$, respectively, while $\tilde{p}_{i,n}$ and $\tilde{\sigma}_{i,n}$ are the model reconstructions. The offset $\epsilon_0{=}0.1$ prevents zero weighting for unreliable detections.  

The robust function $\psi_\delta(\cdot)$ used in our formulation corresponds to the Huber loss~\cite{huberloss}, defined as:
\begin{equation}
\psi_\delta(r) =
\begin{cases}
\dfrac{1}{2\delta}\|r\|_2^2, & \text{if } \|r\|_2 \le \delta,\\[6pt]
\|r\|_2 - \dfrac{\delta}{2}, & \text{otherwise},
\end{cases}
\qquad \delta = 1.
\end{equation}
This loss combines the advantages of $\ell_2$ and $\ell_1$ norms, offering sensitivity to small residuals while reducing the effect of large deviations, which frequently arise from occlusions or motion blur in pose estimation. The smooth transition controlled by $\delta$ stabilizes training under noisy joint detections in normalized coordinate space. Additionally, the weighting ratio between the coordinate reconstruction and confidence sub-terms (set to $0.8$ and $0.2$, respectively) was empirically determined through CV (see Section~\ref{modelvalid}).  Using the Huber loss~\cite{huberloss} in place of a conventional mean-squared error (MSE) function enhances robustness to outliers introduced by partial occlusions, motion blur, or pose-estimation inaccuracies~\cite{girshick2015fast}. Moreover, the proposed confidence-weighted formulation—where each keypoint’s contribution is scaled by its detection reliability $(\sigma_{i,n} + \epsilon_0)$—encourages the model to prioritize well-localized joints while reducing the influence of uncertain or missing detections.


\subsubsection{Emotion Reconstruction Objective}
The reconstruction of facial emotion states is enforced using a Kullback–Leibler (KL) divergence between the ground-truth and reconstructed emotion-probability vectors:
\begin{equation}
\mathcal{J}_{\text{emotion}}
= \frac{1}{T_r}\sum_{i=1}^{T_r}\sum_{c=1}^{7}
g^{\text{emo}}_{i+1,r,c}\,
\log\!\frac{g^{\text{emo}}_{i+1,r,c}}
{\tilde{g}^{\text{emo}}_{i+1,r,c}},
\label{eq:new_affect_loss}
\end{equation}
where $g^{\text{emo}}_{i+1,r,c}$ and $\tilde{g}^{\text{emo}}_{i+1,r,c}$ correspond to the true and predicted emotion probabilities for category $c$. This objective preserves the temporal and probabilistic consistency of emotion dynamics.

\subsubsection{Intent Label Loss}
The reconstruction of interaction intent is formulated as a binary classification task, optimized using a \textit{binary cross-entropy (BCE)} criterion. Since the decoder outputs sigmoid-activated probabilities representing interaction likelihoods, the corresponding loss is defined as 
$\mathcal{J}_{\text{intent}} = \mathrm{BCE}(\ell_{i+1,r}, \hat{\ell}_{i+1,r})$, 
where $\ell_{i+1,r}$ and $\hat{\ell}_{i+1,r}$ denote the ground-truth and predicted frame-level intent labels, respectively.

\subsubsection{KL Latent Regularization with Free-Bits Strategy}

The variational encoder produces a posterior distribution  
\textbf{$q_{\phi}(\mathbf{h}_r|\mathcal{V}_r)=\mathcal{N}(\boldsymbol{\mu}_\phi,\mathrm{diag}(\boldsymbol{\sigma}_\phi^2))$},  
where $\boldsymbol{\mu}_\phi$ and $\boldsymbol{\sigma}_\phi$ denote the mean and standard deviation parameters inferred from the input sequence $\mathcal{V}_r$.  
A standard Gaussian prior \textbf{$p(\mathbf{h}_r)=\mathcal{N}(\mathbf{0},\mathbf{I})$} is imposed to regularize the latent representation.  
For each latent dimension $d$, the  KL divergence between the approximate posterior and the prior is given by:
\begin{equation}
    \mathrm{KL}_d = \tfrac{1}{2}\big(\mu_{\phi,d}^2 + \sigma_{\phi,d}^2 - \log \sigma_{\phi,d}^2 - 1 \big).
\end{equation}

To mitigate posterior collapse, we apply the \textit{free-bits} regularization strategy~\cite{10.5555/3157382.3157627}, which enforces a minimum information content per latent dimension by constraining the KL term to a lower bound $\delta_{\text{FB}}$:
\begin{equation}
\mathcal{J}_{\text{KL}} 
= \sum_{d=1}^{D_h} \max\!\big(\mathrm{KL}_d, \delta_{\text{FB}}\big), 
\qquad \delta_{\text{FB}} = 0.1.
\end{equation}

This formulation prevents inactive latent dimensions from collapsing prematurely toward the prior, thereby preserving an expressive and informative latent space.

During training, the KL regularization weight \textbf{$\eta_h$} is gradually increased according to a linear warm-up schedule across epochs \(e\):
\begin{equation}
\eta_h(e) = \eta_{\max}\cdot 
\min\!\left(\frac{e}{E_{\text{warm}}},\,1\right),
\end{equation}
where \textbf{$\eta_{\max}=0.8$} and \textbf{$E_{\text{warm}}=5000$}.  
This progressive weighting allows the model to focus on accurate reconstruction during early training before applying stronger latent regularization, improving stability and avoiding over-constraint in the initial learning phase.

To stabilize autoregressive decoding, we apply \textit{teacher forcing} and \textit{scheduled sampling}~\cite{Bengio2015ScheduledSampling,WilliamsZipser1989}.  
At each time step \(i\), the decoder input \(\tilde{\mathbf{z}}_i\) is selected as:
\begin{equation}
\tilde{\mathbf{z}}_i =
\begin{cases}
\mathbf{z}_i, & \text{with probability } \tau,\\[4pt]
\hat{\mathbf{z}}_i, & \text{with probability } 1-\tau,
\end{cases}
\end{equation}
where the teacher-forcing probability \(\tau\) is linearly annealed from \(1\) to \(0\) over training epochs.  
This annealing strategy ensures that early iterations benefit from ground-truth conditioning, while later epochs promote autonomous sequence generation, yielding a fully autoregressive decoder at inference.

\begin{figure}[htbp]
    \centering
    \includegraphics[width=0.35\textwidth]{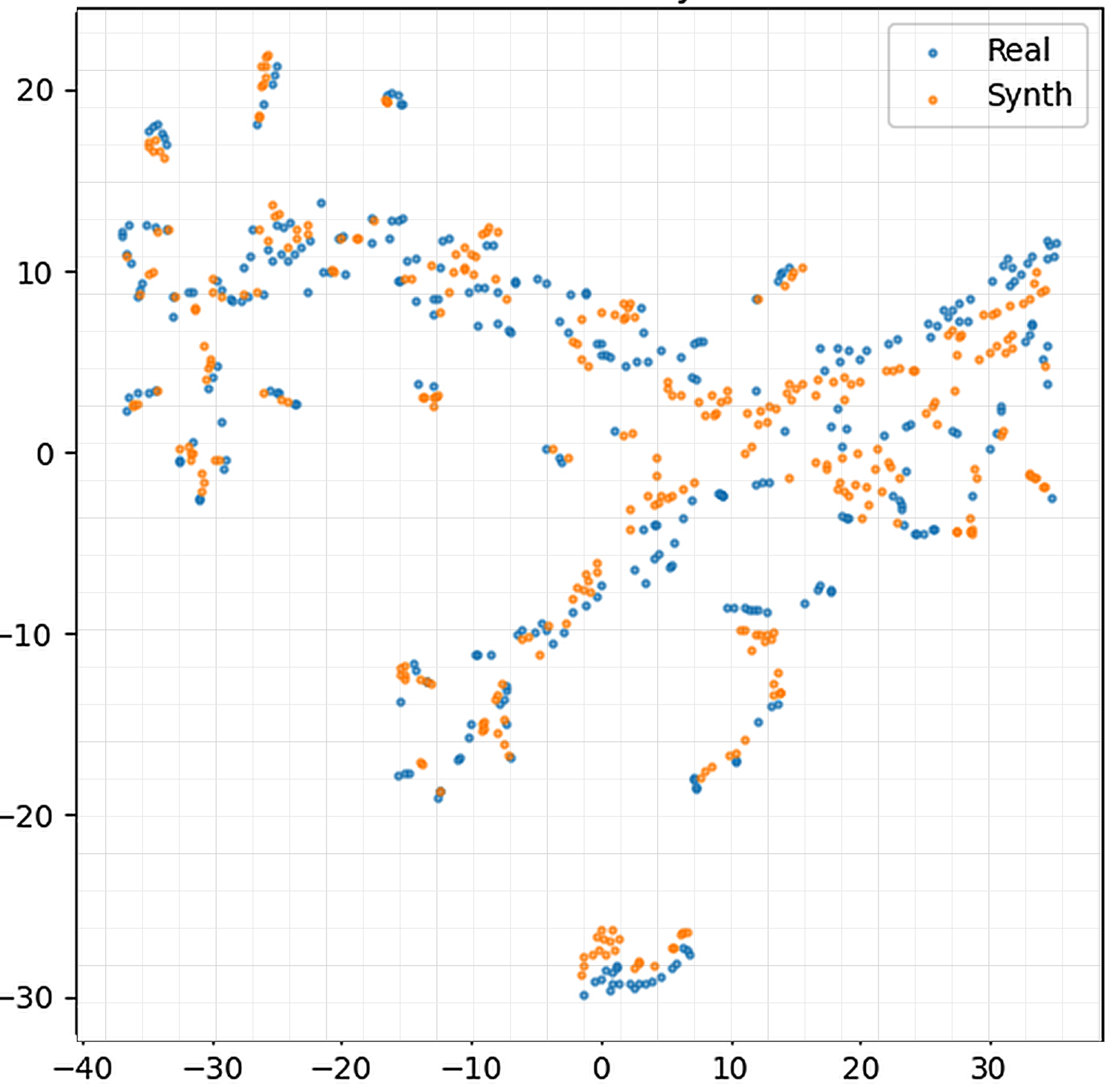}
 \caption{\textit{\textbf{t-SNE visualization of real multimodal data sequences (\textit{Real}) and synthetic embeddings (\textit{Synth}) generated by the proposed MINT-RVAE model.} 
The strong overlap between both distributions demonstrates that MINT-RVAE effectively captures the underlying data manifold, producing synthetic pose–emotion sequences that closely match the real samples.}}
 \vspace{-1em}
    \label{fig:real_vs_synth2}
\end{figure} 
Training is conducted using the Adam optimizer~\cite{kingma2017adammethodstochasticoptimization} with a learning rate of $10^{-3}$, $L_2$ weight decay of $10^{-5}$, and batch size of 64 for 700 epochs.  
The weighting coefficients in the total loss (\ref{eq:mint_total_loss_new}) are empirically set to $\lambda_p=20$, $\lambda_e=10$, and $\lambda_y=1$ to achieve an optimal balance between spatial accuracy, affective reconstruction, and intent-prediction fidelity.  


\subsection{Validation of the MINT-RVAE Generative Model}
\label{modelvalid}

Upon completion of training, latent samples 
$\mathbf{h} \sim \mathcal{N}(\mathbf{0}, \mathbf{I})$ 
are drawn and passed through the autoregressive decoder to synthesize multimodal sequences 
$\{\tilde{\mathbf{g}}^{\text{pose}},\, \tilde{\mathbf{g}}^{\text{emo}},\, \tilde{\ell}\}$ 
representing human motion, affective expression, and intent transitions. 
These generated trajectories are subsequently used to augment the minority intention class during supervised model training.

To quantitatively assess the realism of the synthetic data, we adopt a 
\textit{real-vs-synthetic discrimination protocol} inspired by standard 
time-series generative modeling benchmarks~\cite{yoon2019timegan, pei2021towards}. 
A lightweight recurrent neural network (RNN) is trained on an 80–20\% train–test split 
to classify whether a given sequence originates from the real dataset or from MINT-RVAE. 
We report both the classification accuracy $\mathrm{Acc}$ and the discriminative score:
\begin{equation}
D = \big|\, 0.5 - \mathrm{Accuracy} \,\big|,
\label{eq:discriminative_score}
\end{equation}
where lower $D$ values indicate that the classifier fails to distinguish between real and generated data, implying greater realism of the synthetic sequences. 

The proposed model achieves $\mathrm{Acc} = 0.577$ and $D = 0.077$, 
demonstrating that the generated sequences are highly indistinguishable from real samples. 
Qualitative visualization using t-SNE further supports this finding, 
showing a substantial overlap between the distributions of real and synthesized embeddings as shown in Fig. \ref{fig:real_vs_synth2}




\section{Experiments and Results}
\label{sec:results}





\begin{table*}[t]
\centering


\caption{\textit{\textbf{Cross-subject evaluation results for HRI intent detection.}} Performance of GRU, LSTM, and Transformer backbones under different feature configurations and MINT--RVAE re-balancing.
Metrics are reported at both \textit{frame} and \textit{sequence} levels (mean $\pm$ SD across 5 folds).}

\label{tab:id}
\setlength{\tabcolsep}{3.5pt}
\begin{tabular}{l l c c c c c c}
\toprule
\textbf{Architecture} & \textbf{Variant} & \textbf{Frame F1 (Macro)} & \textbf{Frame Bal. Acc.} & \textbf{Frame AUROC} & \textbf{Seq F1 (Macro)} & \textbf{Seq Bal. Acc.} & \textbf{Seq AUROC} \\
\midrule
\multirow{3}{*}{GRU}
& Pose-only
& 0.814 $\pm$ 0.056 & 0.809 $\pm$ 0.051 & 0.899 $\pm$ 0.066 & 0.845 $\pm$ 0.063 & 0.854 $\pm$ 0.065 & 0.922 $\pm$ 0.042 \\
&   Emotion-only
& 0.466 $\pm$ 0.030 & 0.512 $\pm$ 0.014 & 0.544 $\pm$ 0.057
& 0.250 $\pm$ 0.055 & 0.505 $\pm$ 0.007 & 0.580 $\pm$ 0.155 \\
& Multimodal (no aug)
& 0.845 $\pm$ 0.038 & 0.841 $\pm$ 0.031 & 0.920 $\pm$ 0.031 & 0.867 $\pm$ 0.036 & 0.878 $\pm$ 0.037 & 0.925 $\pm$ 0.025 \\
& \textbf{Multimodal (+VAE)}
& \textbf{0.852 $\pm$ 0.038} & \textbf{0.853 $\pm$ 0.038} & \textbf{0.927 $\pm$ 0.022} & \textbf{0.876 $\pm$ 0.036} & \textbf{0.880 $\pm$ 0.034} & \textbf{0.933 $\pm$ 0.024} \\
\midrule
\multirow{3}{*}{LSTM}
& Pose-only
& 0.835 $\pm$ 0.033 & 0.827 $\pm$ 0.034 & 0.926 $\pm$ 0.019 & 0.869 $\pm$ 0.036 & 0.881 $\pm$ 0.029 & \textbf{0.935 $\pm$ 0.019} \\
& Emotion-only
& 0.407 $\pm$ 0.043 & 0.500 $\pm$ 0.000 & 0.602 $\pm$ 0.013
& 0.544 $\pm$ 0.143 & 0.599 $\pm$ 0.077 & 0.682 $\pm$ 0.030 \\
& Multimodal (no aug)
& 0.838 $\pm$ 0.043 & 0.834 $\pm$ 0.049 & 0.920 $\pm$ 0.026 & 0.874 $\pm$ 0.035 & 0.880 $\pm$ 0.039 & 0.933 $\pm$ 0.024 \\
& \textbf{Multimodal (+VAE)}
& \textbf{0.857 $\pm$ 0.036} & \textbf{0.858 $\pm$ 0.037} & \textbf{0.928 $\pm$ 0.020} & \textbf{0.876 $\pm$ 0.035} & \textbf{0.882 $\pm$ 0.037} & 0.934 $\pm$ 0.022 \\
\midrule
\multirow{3}{*}{Transformer }
& Pose-only
& 0.880 $\pm$ 0.014 & 0.879 $\pm$ 0.014 & 0.947 $\pm$ 0.015 & 0.897 $\pm$ 0.020 & 0.903 $\pm$ 0.018 & 0.948 $\pm$ 0.016 \\
& Emotion-only
& 0.577 $\pm$ 0.043 & 0.589 $\pm$ 0.008 & 0.676 $\pm$ 0.019
& 0.585 $\pm$ 0.050 & 0.636 $\pm$ 0.012 & 0.700 $\pm$ 0.014\\
& Multimodal (no aug)
& 0.881 $\pm$ 0.025 & 0.887 $\pm$ 0.027 & 0.944 $\pm$ 0.018 & 0.897 $\pm$ 0.026 & \textbf{0.908 $\pm$ 0.025} & 0.946 $\pm$ 0.019 \\
& \textbf{Multimodal (+VAE)}
& \textbf{0.888 $\pm$ 0.021} & \textbf{0.895 $\pm$ 0.017} & \textbf{0.95 $\pm$ 0.015} & \textbf{0.899 $\pm$ 0.021} & 0.905 $\pm$ 0.027 & \textbf{0.951 $\pm$ 0.017} \\
\bottomrule
\vspace{-2em}
\end{tabular}
\end{table*}


\subsection{Sequence Modeling}
\label{seq_models}
To model the temporal evolution of human behavior, we train a sequence model that processes feature vectors corresponding to a tracked person over a sliding window of 15 frames and outputs the probability of interaction intent at both the frame and sequence levels. Three temporal backbones are investigated: \texttt{GRU}, \texttt{LSTM}, and a lightweight \texttt{Transformer}. The \texttt{LSTM} and \texttt{GRU} models employ a single recurrent layer followed by a linear classification head. The \texttt{Transformer} variant comprises a compact encoder with learnable positional embeddings, one encoder block, four attention heads, and a LayerNorm–Linear output head. The hidden dimensionality (\(H\)) for each architecture was empirically tuned to achieve optimal
performance during CV experiments.Specifically, \emph{pose-only} models used \(H = 256\), while \emph{emotion-only} models employed \(H = 16\). For the fused \emph{pose+emotion} configuration, multiple settings were explored, with the optimal parameters found to be \(H = 96\) for the \texttt{GRU} and \texttt{LSTM}, and \(H = 256\) for the \texttt{Transformer}. 

Training uses cross-entropy loss and the Adam optimizer~\cite{kingma2017adammethodstochasticoptimization} with a learning rate of \(10^{-3}\) and $l_2$ weight decay of \(10^{-5}\). Each model is trained for 100 epochs with a batch size of 64. All models are trained on a laptop equipped with an NVIDIA RTX~3060 GPU (6\, GB), an Intel Core~i7-11800H CPU, and 16\,GB of RAM.

\subsection{Evaluation Protocols}
\label{eval}
 We evaluate the proposed system under three complementary protocols designed to assess robustness across subjects, environments, and cameras:
 \begin{itemize}
     \item \textit{Cross-Subject:} We perform 5-fold stratified CV on data from Environment~1 and ~2 with single person, ensuring that all frames from each participant remain in the same split. This protocol measures within-scene consistency and generalization to unseen individuals.

     \item \textit{Cross-Scene:} To evaluate generalization across environments and crowd configurations, Environment~1 and ~2 are used for training and validation, while Environment~3 which contains multi-person and dynamic backgrounds is kept entirely unseen during training. We also conduct a two-split repeated held-out experiment by alternating the training subsets of Environment~1 and ~2 to confirm stability.

     \item \textit{On Robot Cross-Camera Deployment:} For real-world deployment evaluation, the best trained model is transferred without fine-tuning to the MIRA robot head, which uses a different onboard RGB camera and operates on a Raspberry~Pi~5 without GPU acceleration (CPU-only). This protocol tests the system’s resilience to domain shifts in imaging characteristics and hardware constraints. For this real-time testing, eight participants were recruited for the real-world testing in the wild within uncontrolled  environment (see Fig. \ref{fig:graphabstr} showcasing the deployment in a restaurant setting). Participants received minimal instructions to ensure naturalistic behavior: they could approach the robot when they intended to interact, or walk past it without interaction intent. The approaching style and general behavior was not imposed and left for the participant to act naturally. Each participant completed four independent trials (32 total), yielding 15 interaction and 17 non-interaction trials, labeled according to the participants’ stated intent. For this real-time evaluation, we compute the confusion matrix and derive accuracy, precision, recall, and \(F_{1}\)-score in Section \ref{realworlddeploy}.
 \end{itemize}

\subsection{Evaluation Metrics}

We report metrics at both the \textit{frame} and \textit{sequence window} levels (i.e., frame-level classification and sequence window-level classification). Frame-level uses per-frame probabilities, while sequence-level makes a decision about the human intention to interact once the model’s predicted intent probability exceeds a threshold $\tau$ for at least $k_{\text{run}}{=}7$ consecutive frames within each 15-frame sequence window (i.e., for half of the frames within each window). For ground-truth sequence labels, a window is positive when the onset annotation yields $\geq 7$ positive frames within the window; otherwise, it is negative.  At the sequence level, true positives are interacting sequences with scores \(\ge \tau\); true negatives are non-interacting sequences with scores \(< \tau\); false positives are non-interacting sequences with scores \(\ge \tau\); and false negatives are interacting sequences with scores \(< \tau\).

We report the area under  AUROC, macro-$F_1$, and balanced accuracy. Macro-$F_1$ (unweighted mean of per-class $F_1$) treats the rare \emph{intent} and common \emph{no-intent} classes equally, making it appropriate under extreme imbalance;  balanced accuracy compensates for skewed class priors and is more interpretable than raw accuracy in our setting; AUROC summarizes separability across all thresholds. All results are reported as \emph{mean $\pm$ standard deviation} across the CV folds.


\begin{figure*}[htbp]
    \setlength{\abovecaptionskip}{0pt}   
    \setlength{\belowcaptionskip}{0pt}   
    \centering
    \includegraphics[width=0.90\textwidth]{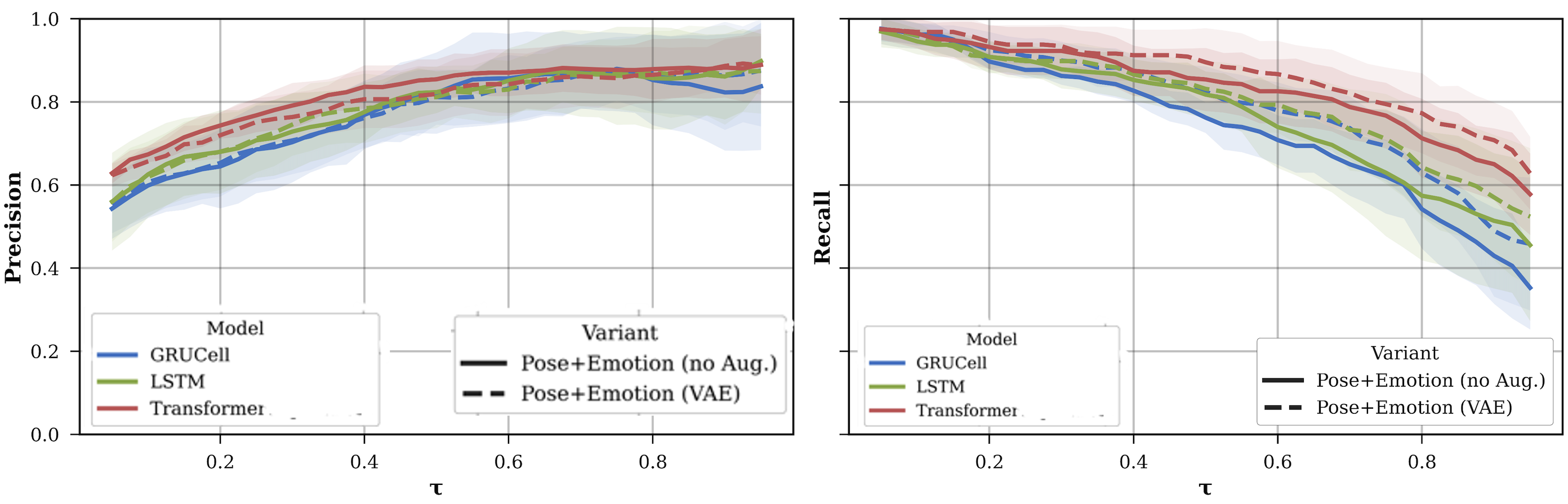}
 \caption{\textit{\textbf{Precision/Recall as a function of the decision threshold $\tau_{ao}$ for the pose+emotion models with and without MINT-RVAE augmentation}.  Mean over 5-fold CV (shaded s.d.). MINT--RVAE shifts the precision–recall trade-off upward.}}
  \vspace{-1em}
    \label{fig:precision-recall}
\end{figure*} 
\subsection{Experimental Results}
This section presents a comprehensive evaluation of the proposed system across offline and real-world settings. Three model architectures are used including (\texttt{GRU}, \texttt{LSTM}, \texttt{Transformer}). Following the protocols described in Section~\ref{eval}, we assess system accuracy, generalization, temporal consistency, and embedded performance. Offline experiments are first conducted under the cross-subject protocol, comparing four configurations: \textit{pose-only}, \textit{emotion-only}, \textit{pose+emotion} without MINT-RVAE augmentation, and the full \textit{pose+emotion} model with MINT-RVAE-based re-balancing. Metrics are reported at both \emph{frame} and \emph{sequence} resolutions. The best-performing configuration is then used for further evaluation under the cross-scene and cross-camera protocols to test robustness to environmental variation and on-robot deployment.


\subsubsection{Cross-Subject Evaluation}

Table~\ref{tab:id} summarizes the results across all backbones and configurations. Three consistent patterns are observed:  First, multimodal fusion of pose and emotion features outperforms unimodal variants, confirming that affective cues provide complementary information to body dynamics for early intention inference. Second, incorporating MINT-RVAE augmentation improves all performance indicators by mitigating sequence-level imbalance and encouraging smoother temporal transitions. Third, the \emph{Transformer} architecture achieves the highest overall results, highlighting the advantage of attention-based temporal modeling in capturing pre-onset behavioral cues.

\begin{figure*}[]
    \centering
    \begin{subfigure}[t]{0.33\linewidth}
        \centering
        \includegraphics[width=\linewidth]{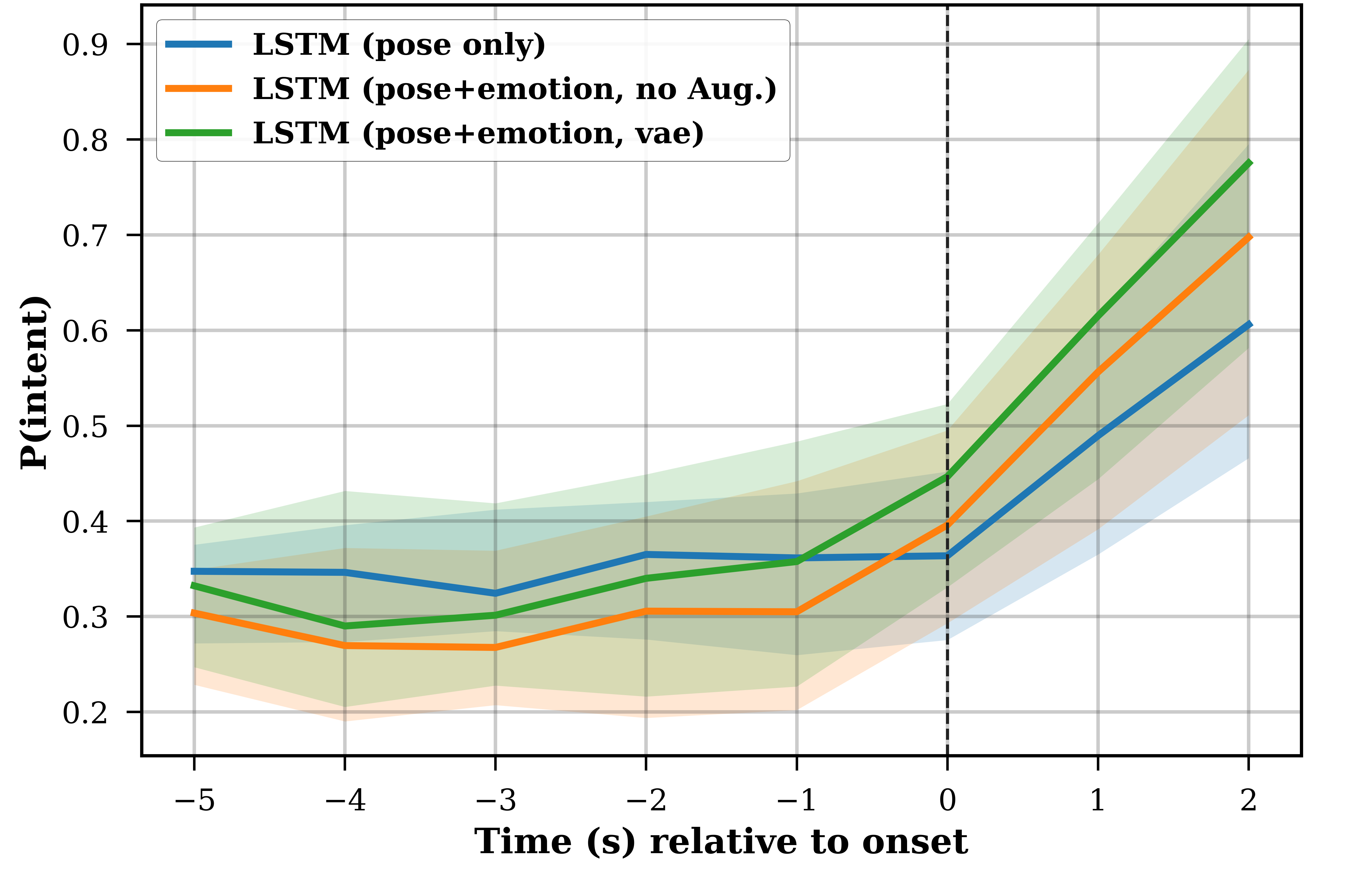}
        \caption{LSTM}
    \end{subfigure}
    \hspace{0.001\linewidth}
    \begin{subfigure}[t]{0.32\linewidth}
        \centering
        \includegraphics[width=\linewidth]{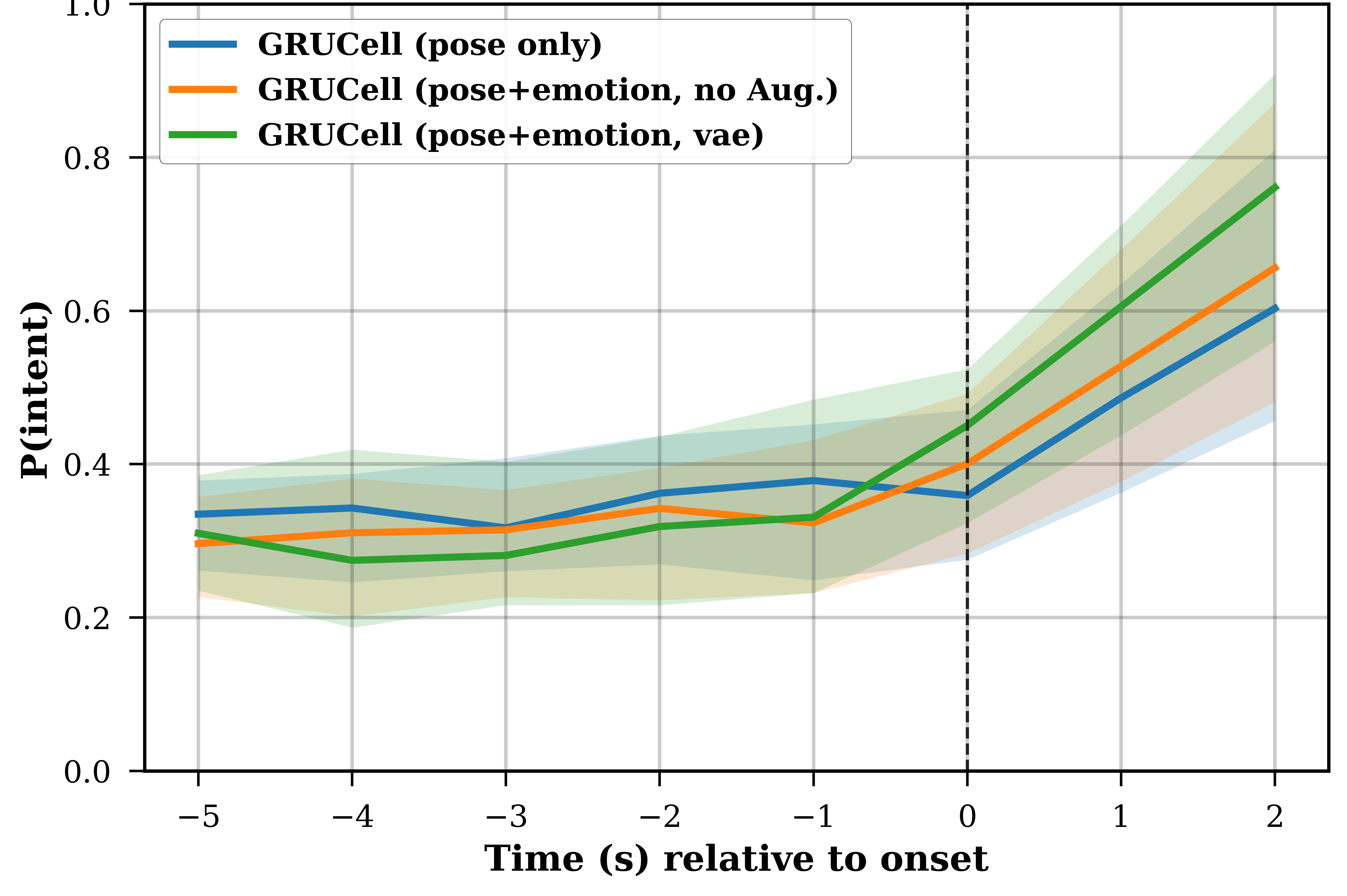}
        \caption{GRU}
    \end{subfigure}
    \hspace{0.001\linewidth}
    \begin{subfigure}[t]{0.32\linewidth}
        \centering
        \includegraphics[width=\linewidth]{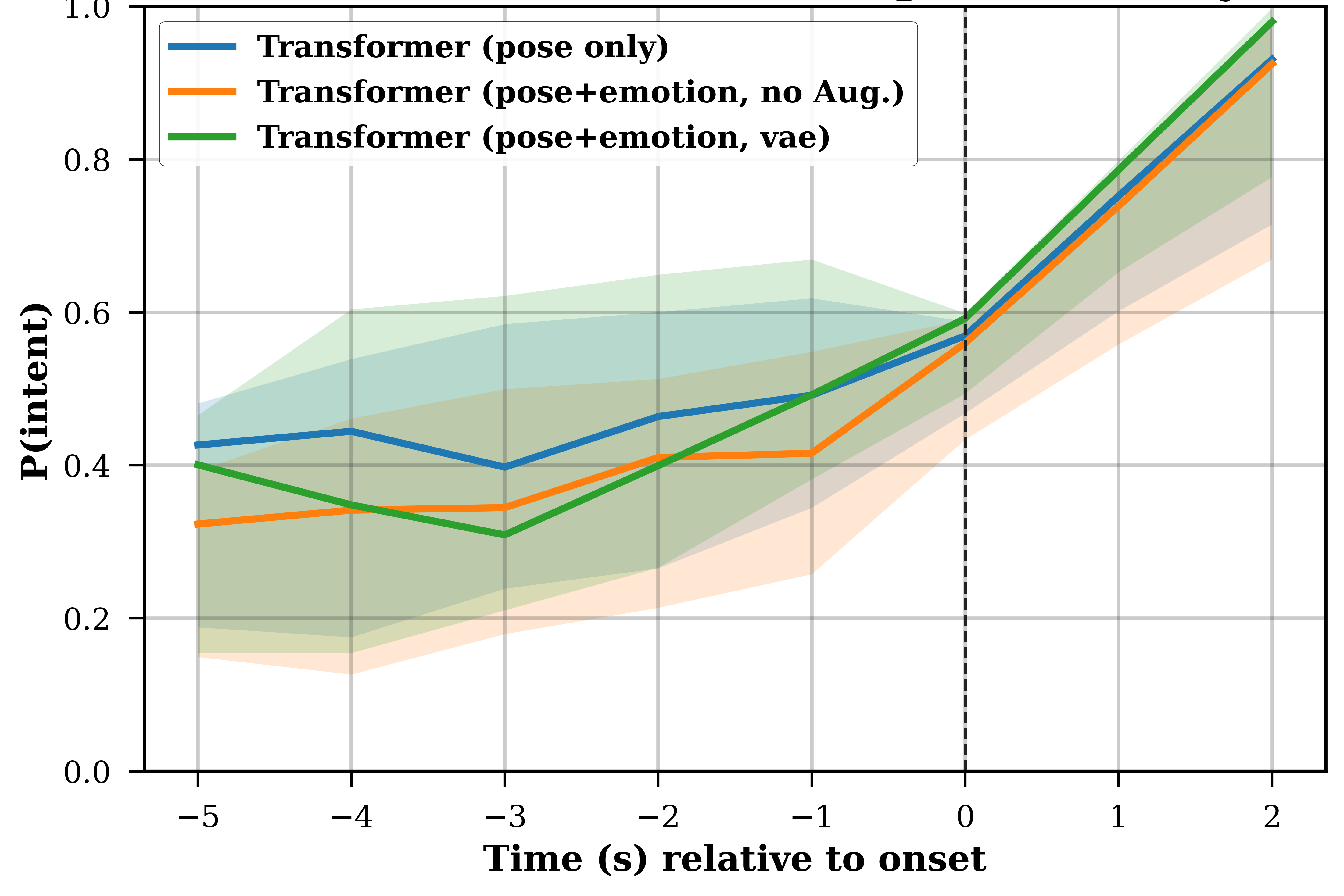}
        \caption{Transformer}
    \end{subfigure}
\caption{\textit{\textbf{Median intent probability trajectories aligned to ground-truth onset ($t=0$).} MINT-RVAE steepens the pre-onset rise, enabling earlier and more reliable detection.}}
    \vspace{-1em}
    \label{fig:prob_vs_time}
\end{figure*}

\textit{At the frame level}, the inclusion of MINT-RVAE in the multimodal configuration improves the macro-F1 score by approximately 4\% on average and increases the AUROC by about 1\% compared to the pose-only baselines, indicating enhanced discrimination of early intent cues. Emotion-only inputs, by contrast, perform substantially worse, indicating that facial affect alone is insufficient for reliable early detection, whereas whole-body motion provides the dominant predictive signal. \textit{At the sequence level}, the multimodal \emph{Transformer} with MINT-RVAE attains the best performance (AUROC of 0.951), reflecting effective aggregation of temporal evidence and more confident intent classification over time. Precision–recall characteristics in Fig.~\ref{fig:precision-recall} further show that MINT-RVAE systematically shifts the trade-off upward, enabling higher recall at comparable precision—particularly for Transformer and GRU models, thus capturing more true interaction events without increasing false alarms. Furthermore, temporal probability trajectories in Fig.~\ref{fig:prob_vs_time} illustrate the same effect dynamically: predicted intent scores rise steadily as the interaction onset approaches ($t{=}0$), with MINT-RVAE-enhanced variants showing earlier separation from baselines and a steeper pre-onset increase. These findings collectively demonstrate that combining multimodal fusion with imbalance-aware temporal augmentation enhances model robustness and consistency under subject-level variability.

\begin{figure}[t]
    \centering
    \begin{subfigure}[t]{0.95\linewidth}
        \centering
        \includegraphics[width=\linewidth]{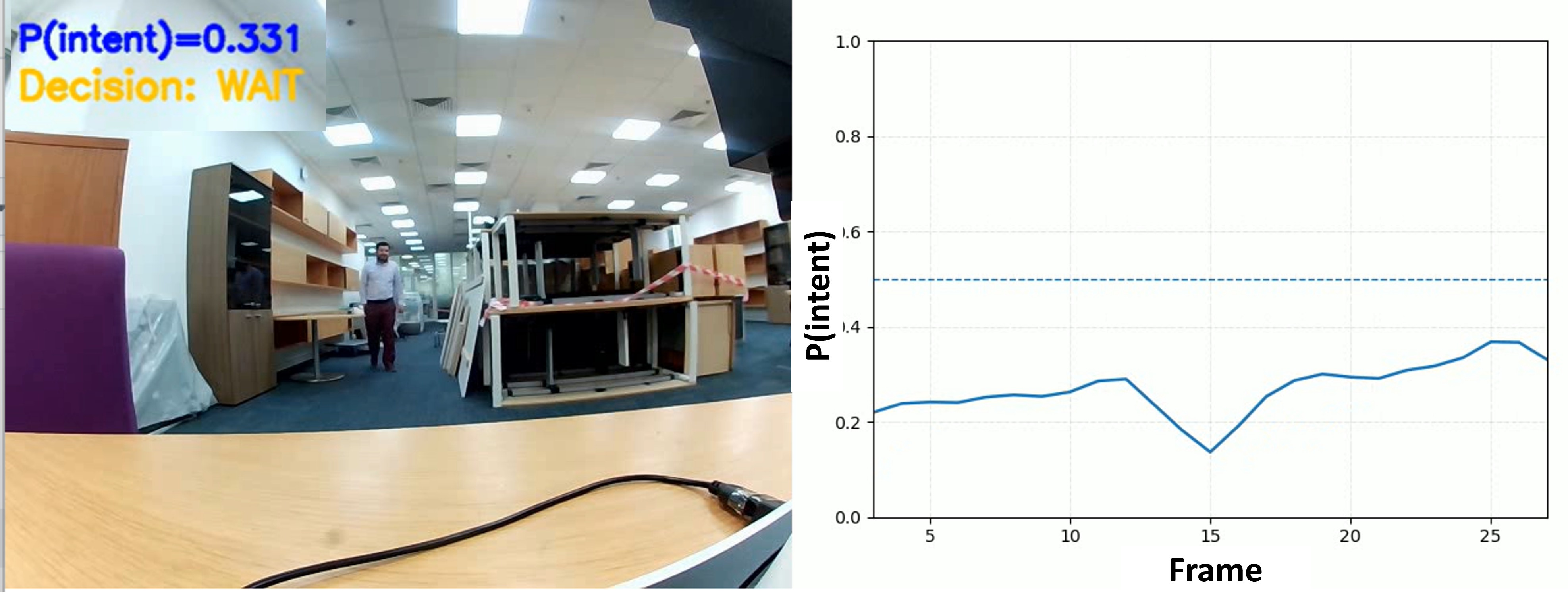}
        \caption{No Intent}
    \end{subfigure}
    \hfill
    \hfill
    \begin{subfigure}[t]{0.95\linewidth}
        \centering
        \includegraphics[width=\linewidth]{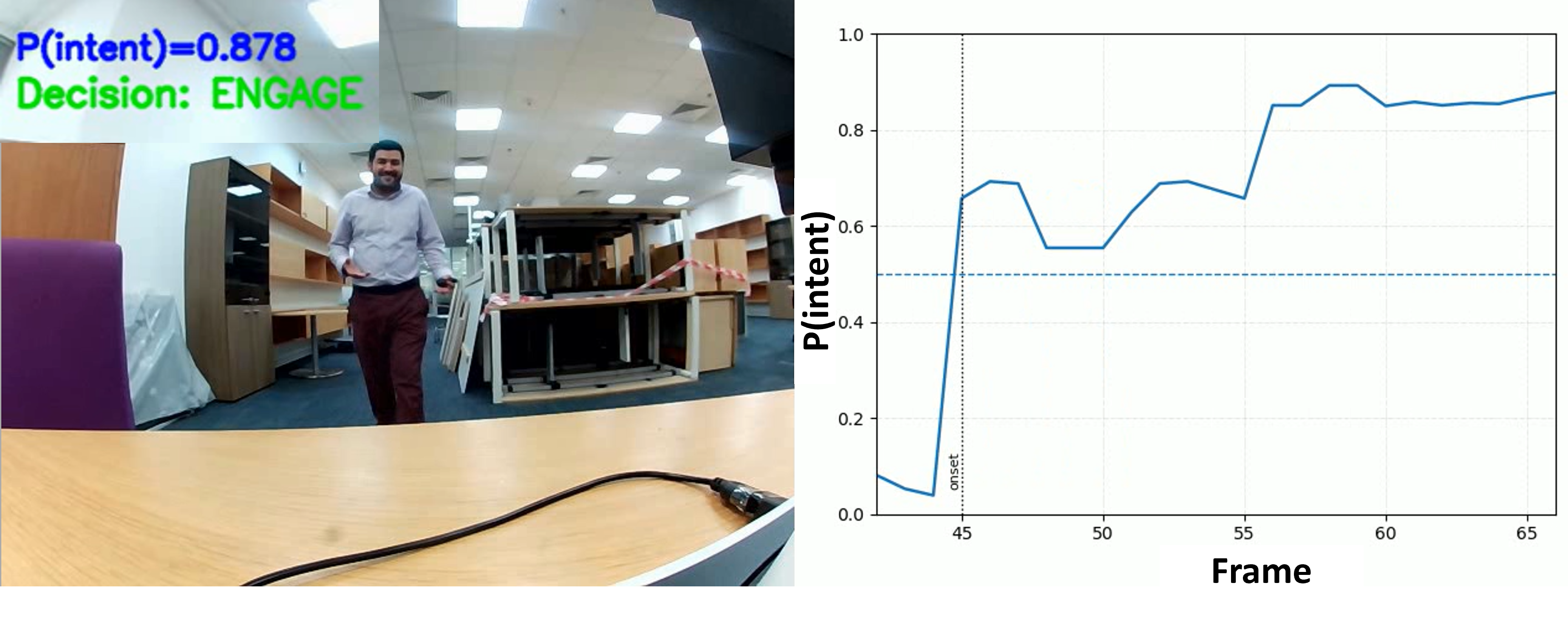}
        \caption{High Intent}
    \end{subfigure}
    \hfill
    \begin{subfigure}[t]{0.95\linewidth}
        \centering
        \includegraphics[width=\linewidth]{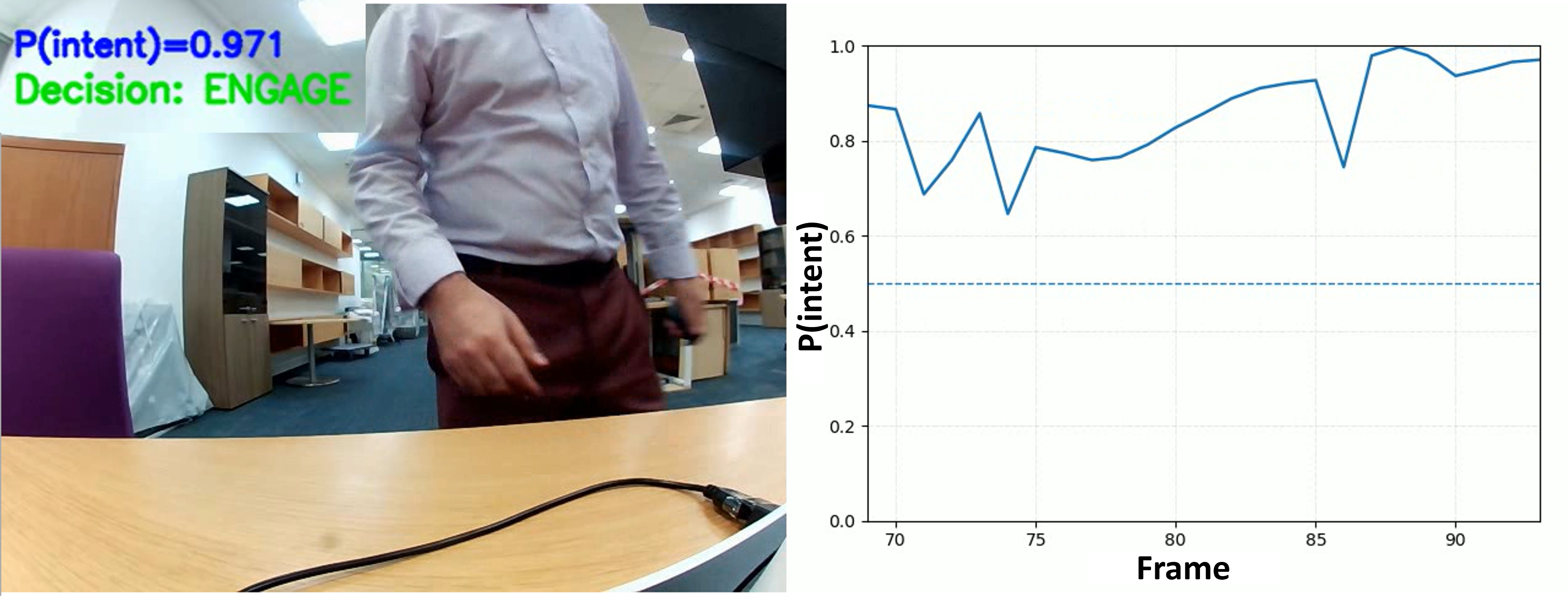}
        \caption{Confirmed Engagement}
    \end{subfigure}

    \caption{\textit{\textbf{Qualitative results}. Each subfigure shows a representative frame (left) and the corresponding temporal evolution of the predicted intent probability (right), generated by the Transformer trained with MINT-RVAE rebalancing. The examples span different stages of approach and interaction, from non-interactive (a) to confirmed interaction (c). The predicted probability rises smoothly across frames and reaches high confidence several frames before the onset of physical interaction.}}
     \vspace{-1em}
    \label{fig:mira_qualitative}
    \vspace{-1em}
\end{figure}

Qualitative deployment examples are shown in Fig.~\ref{fig:mira_qualitative}, illustrating the temporal evolution of intent probabilities during representative user approaches. Using the Transformer model with MINT-RVAE re-balancing, as the person enters the robot’s social space, the predicted intent probability rises smoothly, crossing the engagement threshold several frames before physical interaction occurs. Non-interactive sequences (\textit{WAIT}) remain near baseline, while interactive cases (\textit{ENGAGE}) show stable and monotonic probability growth. These results confirm that the system produces temporally coherent predictions, validating its robustness for field-deployed human–robot interaction.

\subsubsection{Cross-Scene Evaluation}

The cross-scene protocol examines out-of-distribution generalization to unseen environments. Models are trained exclusively on single-person data from Environments 1–2 and evaluated on the held-out two-person set. This setup introduces both visual and behavioral domain shifts, as no data from Environment 3 is seen during training. As shown in Table \ref{tab:env3_results}, incorporating the proposed MINT-RVAE re-balancing strategy during training yields a marked improvement in generalization across all architectures, with  an average multiplicative gain of $\sim 25\%$ across all the performance indicators. Among all evaluated models, the Transformer + MINT-RVAE configuration achieves the best results, attaining AUROC = 0.932 at the frame level and 0.957 at the sequence level. Substantial performance gains are also observed for the GRU and LSTM variants, confirming that our MINT-RVAE generalizes effectively even in complex, multi-person interactions. 



\begin{table*}[t]
\centering
\caption{\textit{\textbf{Cross-Scene evaluation on out-of-sample multi-person test data (Environment 3).} The performance is reported as the \texttt{mean} $\pm$ the \texttt{standard deviation} across the different training data folds (keeping the whole of Env3 as the test set during the evaluation procedure). Best values are indicated in bold.}}
\label{tab:env3_results}
\setlength{\tabcolsep}{3.6pt}
\begin{tabular}{l l c c c c c c}
\toprule
\textbf{Architecture} & \textbf{Variant} &
\textbf{Frame F1 (Macro)} &
\textbf{Frame Bal. Acc.} & \textbf{Frame AUROC} &
\textbf{Seq F1 (Macro)} & \textbf{Seq Bal. Acc.} & \textbf{Seq AUROC} \\
\midrule
\multirow{2}{*}{GRU}
& Multimodal (no aug) 
& 0.497 $\pm$ 0.075 & 0.540 $\pm$ 0.042 & 0.575 $\pm$ 0.247 
& 0.459 $\pm$ 0.322 & 0.610 $\pm$ 0.436 & 0.636 $\pm$ 0.361 \\
& \textbf{Multimodal (+VAE)} 
& \textbf{0.728 $\pm$ 0.020} & \textbf{0.723 $\pm$ 0.006} & \textbf{0.770 $\pm$ 0.007} 
& \textbf{0.636 $\pm$ 0.063} & \textbf{0.859 $\pm$ 0.044} & \textbf{0.810 $\pm$ 0.054} \\
\midrule

\multirow{2}{*}{LSTM}
& Multimodal (no aug) & 0.508 $\pm$ 0.130 & 0.554 $\pm$ 0.060 & 0.525 $\pm$ 0.101 & 0.425 $\pm$ 0.246 & 0.567 $\pm$ 0.084 & 0.579 $\pm$ 0.198 \\
& \textbf{Multimodal (+VAE)} & \textbf{0.742 $\pm$ 0.032} & \textbf{0.734 $\pm$ 0.026} & \textbf{0.820 $\pm$ 0.059} & \textbf{0.781 $\pm$ 0.024} & \textbf{0.810 $\pm$ 0.024} & \textbf{0.856 $\pm$ 0.026} \\
\midrule

\multirow{2}{*}{Transformer}
& Multimodal (no aug) 
& 0.749 $\pm$ 0.112 & 0.735 $\pm$ 0.097 & 0.856 $\pm$ 0.113 
& 0.812 $\pm$ 0.089 & 0.824 $\pm$ 0.036 & 0.876 $\pm$ 0.101 \\
& \textbf{Multimodal (+VAE)} 
& \textbf{0.820 $\pm$ 0.012} & \textbf{0.809 $\pm$ 0.023} & \textbf{0.932 $\pm$ 0.007} 
& \textbf{0.862 $\pm$ 0.023} & \textbf{0.869 $\pm$ 0.013} & \textbf{0.957 $\pm$ 0.002} \\

\bottomrule
\end{tabular}
\end{table*}

Compared with the cross-subject setting, overall accuracy decreases slightly—on average by approximately 3–5\%—reflecting the increased challenge of generalizing to unseen backgrounds, lighting, and multi-person dynamics. However, models trained with the proposed MINT-RVAE exhibit markedly smaller performance degradation, demonstrating enhanced robustness. Notably, the \emph{Transformer} with MINT-RVAE maintains near-identical performance across scenes, underscoring its resilience and capacity to generalize effectively to out-of-distribution conditions.

\color{black}

\subsubsection{On Robot Cross-Camera Deployment}
\label{realworlddeploy}

To evaluate real-world deployment performance, the best-performing model from the offline experiments (multimodal \textit{pose+emotion} with MINT-RVAE re-balancing and \emph{Transformer} backbone) was deployed without fine-tuning on the MIRA robot’s onboard RGB camera, following the protocol described in Section~\ref{eval}. The complete inference pipeline including person detection, 2D pose estimation, face emotion extraction, and intention prediction is executed fully on-device using a Raspberry~Pi~5 CPU, without GPU acceleration or external computation. This deployment setting differs substantially from the training environment in both camera viewpoint and imaging characteristics, thus constituting a rigorous test of cross-camera and cross-environment generalization under embedded hardware constraints.

\begin{figure}[]
\centering
\includegraphics[width=0.6\columnwidth]{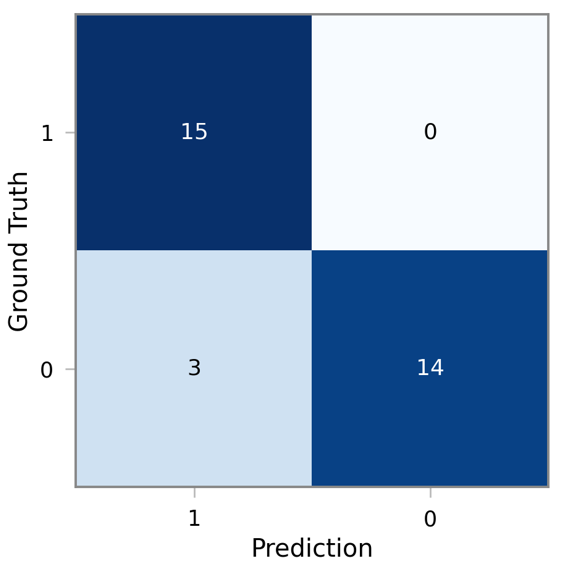}
\caption{\textit{\textbf{Confusion matrix for real-world deployment}. The system achieved 100\% recall (15/15 interaction trials detected) with three false alarms during non-interaction scenarios.}}
 \vspace{-1em}
\label{fig:confusion_matrix}
\end{figure}

Quantitatively, the deployed system demonstrates strong and reliable performance in uncontrolled real-world conditions.
As summarized in Fig.~\ref{fig:confusion_matrix} and Table~\ref{tab:deployment_metrics}, the model achieves perfect recall for interaction intent by successfully detecting all 15 positive trials (true positives = 15, false negatives = 0), and correctly identifying 14 out of 17 non-interaction trials (true negatives = 14, false positives = 3).
Achieving perfect recall is particularly valuable in service robotics, as missed engagement cues can directly impair user experience.
The overall accuracy and F$_1$-score reach 0.91, with a precision of 0.83, confirming robust, real-time intent recognition across unseen participants and spontaneous behaviors. Despite being trained on USB webcam data and deployed on the robot’s onboard RGB camera with different resolution and optics, the system maintains high accuracy thanks to the bounding-box normalization of 2D pose coordinates, which yields camera-invariant representations. This normalization ensures scale and translation invariance, allowing the model to capture relative pose dynamics without relying on explicit 3D geometry and demonstrating that RGB-only sensing is useful for early intention detection. \textit{Importantly}, on-board execution on a CPU-only Raspberry~Pi~5 confirms the feasibility of real-time intent recognition on low-cost social robots such as MIRA, reducing sensing and compute requirements while maintaining competitive accuracy.

\begin{figure*}[t]
    \centering
    
        \includegraphics[width=\textwidth]{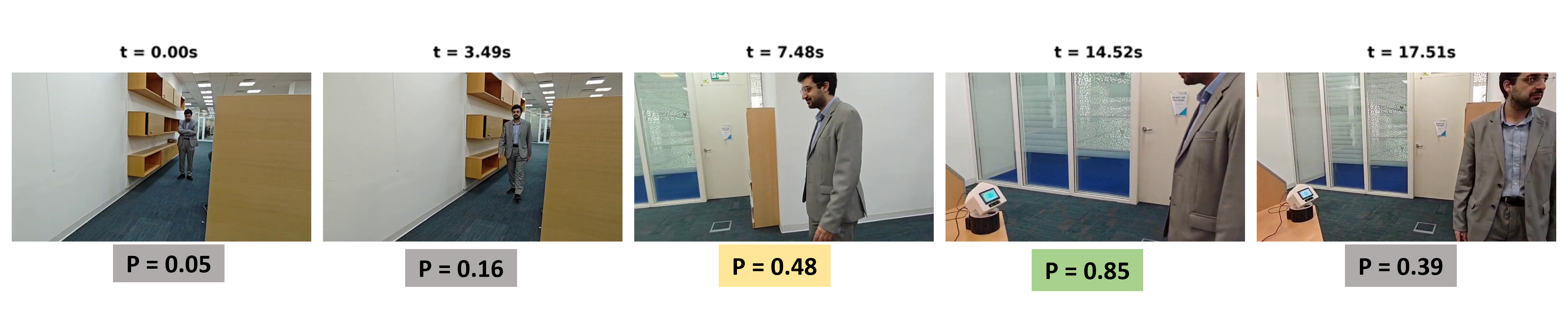}
  

\caption{\textbf{Real-world deployment demonstrating dynamic engagement and disengagement detection.}
\textit{Shown:} five representative frames from a 22-second on-robot trial illustrating the evolution of interaction intent. The predicted intent probability ($P$) rises gradually during approach ($t = 0$--$7.5~\mathrm{s}$, $P = 0.05 \rightarrow 0.48$), peaks at engagement ($t = 14.5~\mathrm{s}$, $P = 0.85$) when the participant faces the robot—triggering the green display and greeting response—and declines upon disengagement ($t = 17.5~\mathrm{s}$, $P = 0.39$) as the participant turns away. Color-coded boxes indicate predicted probability levels (gray: low intent, yellow: transitional, green: intention detected).}
\label{fig:deployment_scene}
 
\end{figure*}

Figure~\ref{fig:deployment_scene} illustrates a representative 22-second on-robot deployment trial capturing both interaction and non-interaction phases.
The predicted intent probability ($P$) rises steadily from a low baseline ($P = 0.05$) as the participant approaches the robot, reaching a peak of $P = 0.85$ at approximately $t = 14.5~\mathrm{s}$ when the participant directly faces the robot.
At this point, the robot's display transitions to green and triggers a greeting response, marking the onset of engagement detection.
Subsequently, as the participant turns away, the intent probability decreases to $P = 0.39$, and the robot display reverts to its neutral state, signaling no intent to interact. This continuous and smooth evolution of predicted intent demonstrates the model's temporal stability and its sensitivity to social cues such as body orientation and head pose, enabling context-aware and socially appropriate robot behavior~\cite{satake2009approach,dautenhahn2007socially}.
This video, along with additional deployment examples, is provided in the supplementary material (Videos~S1--S3), showcasing successful intent detection, non-interaction, and combined engagement--disengagement scenarios.


\begin{table}[t]
\centering
\caption{\textit{\textbf{Real-world deployment performance on MIRA robot.}}}
\label{tab:deployment_metrics}
\setlength{\tabcolsep}{8pt}
\begin{tabular}{cccc}
\toprule
Accuracy & Precision & Recall & F$_1$-score \\
\midrule
0.91 & 0.83 & 1.00 & 0.91 \\
\bottomrule
\end{tabular}
\vspace{-2em}
\end{table}

\subsection{Task Adaptability}

Although the proposed framework is validated across specific interaction scenarios, its underlying design supports a high degree of task adaptability. The model does not rely on any task-dependent features, cues, or priors; instead, it learns spatiotemporal intent patterns from body motion and facial emotion signals that are broadly transferable across social contexts. Both the training and evaluation datasets comprise spontaneous, unstructured human behaviors that are not constrained by predefined robot tasks, demonstrating that the system generalizes to natural variations in human approach, posture, and expression. The use of camera-invariant, bounding-box–normalized pose representations ensures that spatial reasoning depends on relative geometry rather than fixed scene coordinates, allowing deployment across different viewpoints and robot configurations. Similarly, the MINT-RVAE augmentation synthesizes temporally coherent intent sequences independent of task semantics, reinforcing the model’s focus on temporal dynamics rather than scene-specific patterns. The deployment experiments further confirm this property: the model trained on passive office interactions successfully transfers to active service-robot settings without fine-tuning, despite differences in task context, background, and participant behavior. Collectively, these findings support that the proposed method captures task-agnostic behavioral cues indicative of human engagement, making it suitable for diverse human–robot interaction tasks such as greeting, assistance, or attention monitoring without additional retraining.

\subsection{Comparison with Previous Works}
We contextualize our proposed \textit{RGB-only} human–intention detection framework within recent research in HRI.
Table~\ref{tab:comparison} summarizes representative intent-prediction systems, emphasizing differences in sensing modality, temporal granularity, imbalance handling, and deployment feasibility. Most prior approaches rely on \textit{RGB–D} or multimodal inputs combining depth, gaze, or audio cues~\cite{abbate2024self,Arreghini2024,trick2023can,liu2019rgbd,zimmermann2018pose}. While these systems report strong accuracy, they typically depend on specialized RGB–D sensors (e.g., Kinect, RealSense) and GPU-based inference, which may limit their affordability and practicality to  low-resource social robots. In contrast, our framework operates solely on monocular RGB input and achieves superior  offline performance (AUROC $= 0.951 \pm 0.017$ in 5-fold CV using GPU inference). Moreover, after offline training, the same Transformer-based model was deployed without fine-tuning—on a \textit{CPU-only Raspberry~Pi~5} onboard the MIRA robot that has different camera characteristics, achieving 91\% accuracy and perfect recall in real-world trials.

Another distinction lies in how dataset imbalance is handled.
Most existing systems do not explicitly address the strong class imbalance characteristic of natural HRI datasets, where positive interaction instances are rare~\cite{thompson2024pard,natori2025mall}.
Our method incorporates MINT-RVAE–based sequence augmentation, which synthesizes temporally coherent pose–emotion data for underrepresented intent-onset events. This strategy improves the model’s sensitivity to rare classes and leads to more balanced decision boundaries across test domains. Moreover, previous studies primarily evaluate at the \textit{sequence} or \textit{event} level, where all frames within a clip share the same label~\cite{Abbate2024,Arreghini2024}, thus limiting temporal precision and anticipatory responsiveness. Our approach instead performs frame-level inference.

\begin{table*}[t]
\centering
\footnotesize
\setlength{\tabcolsep}{3pt}
\renewcommand{\arraystretch}{1.05}

\caption{\textit{\textbf{Comparison with representative intent-prediction methods in HRI}}
Reported works differ in sensing modality, temporal resolution, class-imbalance treatment, and deployment feasibility.
Our method achieves competitive offline performance (GPU-based) and validated real-time deployment on CPU-only embedded hardware.}

\label{tab:comparison}
\resizebox{\textwidth}{!}{%
\begin{tabularx}{\textwidth}{@{}
>{\raggedright\arraybackslash}p{2.9cm}
>{\raggedright\arraybackslash}p{2.3cm}
>{\centering\arraybackslash}p{2.2cm}
>{\raggedright\arraybackslash}p{2.7cm}
>{\centering\arraybackslash}p{1.6cm}
>{\raggedright\arraybackslash}p{2.6cm}
>{\raggedright\arraybackslash}p{3.1cm}
@{}}
\toprule
Work & Sensors & Temporal Resolution & Features & Data Re-balancing & Deployment Platform & Reported Performance \\
\midrule
Zimmermann et al.\ (2018)~\cite{zimmermann2018pose} 
& RGB--D 
& Sequence-level 
& Pose, gesture 
& No 
& GPU workstation 
& Accuracy $\approx 0.87$ \\

Liu et al.\ (2019)~\cite{liu2019rgbd} 
& RGB--D 
& Sequence-level 
& 3D skeleton 
& No 
& GPU workstation 
& F$_1$ $\approx 0.84$ \\

Trick et al.\ (2023)~\cite{trick2023can} 
& RGB--D + audio 
& Sequence-level 
& Pose, gaze, speech 
& No 
& GPU workstation 
& F$_1$ $\approx 0.81$ \\

Abbate et al.\ (2024)~\cite{abbate2024self} 
& RGB--D 
& Sequence-level 
& Pose (self-supervised) 
& No 
& GPU workstation 
& AUROC $\approx 0.90$ \\

Arreghini et al.\ (2024)~\cite{Arreghini2024} 
& RGB--D + gaze 
& Sequence-level 
& Pose, gaze cues 
& No 
& GPU workstation 
& AUROC $= 0.912$ \\

Kedia et al.\ (2024)~\cite{Kedia2024} 
& RGB--D 
& Sequence-level 
& Pose, robot action 
& No 
& GPU workstation 
& Accuracy $\approx 0.89$ \\

\midrule
\textbf{Ours (MINT-RVAE)} 
& \textbf{RGB-only} 
& \textbf{Frame + Sequence-level} 
& \textbf{Pose + Emotion} 
& \textbf{Yes (MINT-RVAE)} 
& \textbf{CPU-only (Raspberry~Pi~5)} 
& \textbf{AUROC: \hspace{15pt} $=0.951\!\pm\!0.017$ (CV), Accuracy $=91\%$ (real-world)} \\

\bottomrule
\end{tabularx}%
}
\end{table*}

\section{Conclusion}
\label{sec:conclusion}

This paper presented a practical, deployable framework for frame-accurate HRI detection using only monocular RGB input. Unlike prior RGB-D or multimodal systems that depend on costly hardware and GPU acceleration, the proposed method achieves superior accuracy and generalization while operating on low-power, CPU-only, and different camera robotic head. By integrating camera-invariant 2D pose and facial-emotion features within a temporal modeling architecture, the system enables early and reliable recognition of human interaction intent without relying on depth information, thereby preserving affordability. To address the severe class imbalance inherent in natural HRI data, we introduced MINT-RVAE, which generates temporally coherent, label-consistent pose–emotion sequences for effective data re-balancing.
This generative augmentation was shown to enhance the discriminative power of temporal models (GRU, LSTM, Transformer) and to improve robustness across under cross-subject, cross-scene, and cross-camera evaluation protocols. Empirically, the approach achieved an AUROC of 0.951 in cross-validation and maintained 0.91 accuracy during real-world, CPU-only deployment on the MIRA robot, validating both its algorithmic robustness and practical embedded feasibility. Beyond its quantitative performance, the proposed pipeline demonstrates that effective social-interaction understanding can be realized without depth sensing, heavy computation, or task-specific supervision. These findings establish a new direction for resource-aware multimodal perception in service robotics, extending the reach of HRI intention detection to low-cost, real-world deployments.




%




\ifCLASSOPTIONcaptionsoff
  \newpage
\fi

\bibliographystyle{IEEEtran}
\bibliography{refs}

\begin{IEEEbiography}[{\includegraphics[width=1in,height=1.5in,clip,keepaspectratio]{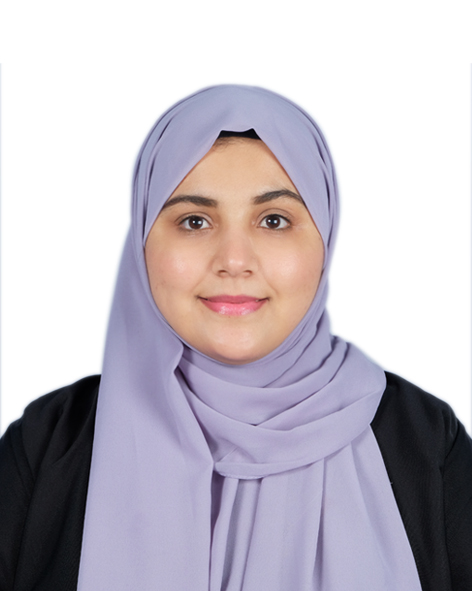}}]{Farida Mohsen}
is currently a Postdoctoral Researcher with the College of Science and Engineering, Hamad Bin Khalifa University (HBKU), Doha, Qatar. 
She received the Ph.D. degree in Computer Science and Engineering from the College of Science and Engineering, HBKU, Doha, Qatar, the M.S. degree in Computer Science from Central South University, China, and the B.S. degree in Information Technology and Computer Science from Cairo University, Egypt. She has authored and co-authored more than 27 research articles in international journals and conferences. 
Her research interests include human–robot interaction, social robotics, edge AI, sensor fusion, medical AI, multimodal data integration, and natural language processing.
\end{IEEEbiography}

\begin{IEEEbiography}[{\includegraphics[width=1in,height=1.5in,clip,keepaspectratio]{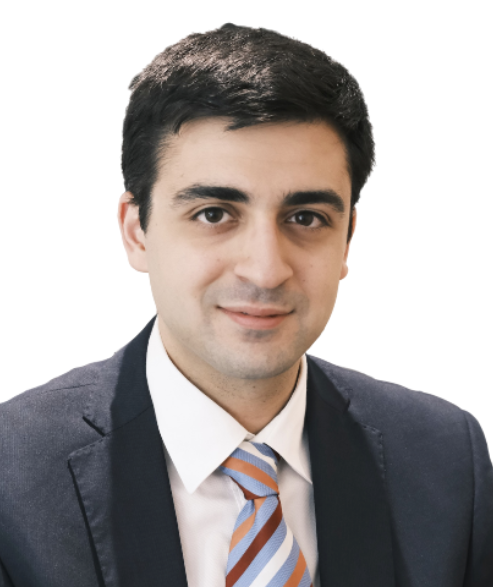}}]{Ali Safa} 
(Member, IEEE) is currently an Assistant Professor with the College of Science and Engineering, Hamad Bin Khalifa University (HBKU), Doha, Qatar. He received the M.Sc. degree in electrical engineering from the Université Libre de Bruxelles, Brussels, Belgium, and the Ph.D. degree in AI-driven processing for robotics and extreme edge applications from the KU Leuven, Belgium, in collaboration with the IMEC R\&D Center, Leuven Belgium. He has been a Visiting Researcher with the University of California at San Diego (UC San Diego), La Jolla, USA, in Spring 2023. He has authored and co-authored more than 50 research articles in international journals and conferences, and is the author of a Springer book on the application of Neuromorphic Computing to Sensor Fusion and Drone Navigation tasks. His research interests include human-robot interaction, social robotics, Edge AI, continual learning, and sensor fusion for robot perception. 
\end{IEEEbiography}

\end{document}